%% file: main.tex
\pdfoutput=1
\documentclass[11pt]{article}

\usepackage[final]{acl}
\usepackage{times}
\usepackage{latexsym}
\usepackage{amsmath}
\usepackage{enumitem}
\usepackage{subcaption} % Paquete para subfiguras
\usepackage[T1]{fontenc}
\usepackage[utf8]{inputenc}
\usepackage{siunitx}
\usepackage{microtype}

\usepackage{inconsolata}

\usepackage{graphicx}
\usepackage{natbib}
\usepackage{mathtools}

\usepackage{booktabs}
\usepackage{xcolor}
\usepackage{colortbl}
\usepackage{amssymb}
\usepackage{stfloats}

\usepackage{makecell}
\usepackage{enumitem}
\usepackage{multirow}
\usepackage{makecell}

\usepackage{colortbl}  % Para colorear tablas
\usepackage{xcolor}    % Para definir escalas de color
\usepackage{booktabs}

\usepackage{array}
\usepackage{float}
\usepackage{caption}
\usepackage{enumitem}

\usepackage{float}       % Para usar [H] en las tablas
\usepackage{tabularx}
\usepackage{booktabs}
\renewcommand{\arraystretch}{1.25}
\usepackage{caption}     % Opcional: para estilos de caption si usas alguno

\definecolor{VeryLowColor}{RGB}{255,150,150}   % Nivel 1: Muy bajo
\definecolor{LowColor}{RGB}{255,200,200}       % Nivel 2: Bajo (usado también para Δ negativas)
\definecolor{MediumColor}{RGB}{255,255,200}    % Nivel 3: Medio
\definecolor{HighColor}{RGB}{200,255,200}        % Nivel 4: Alto (usado también para Δ positivas)
\definecolor{VeryHighColor}{RGB}{150,255,150}    % Nivel 5: Muy alto

\newcommand{\mycomment}[1]{}

\title{The Reader is the Metric: How Textual Features and Reader Profiles Explain Conflicting Evaluations of AI Creative Writing}
\author{
 \textbf{Guillermo Marco\textsuperscript{\dag}},
 \textbf{Julio Gonzalo\textsuperscript{\dag}},
  \textbf{Víctor Fresno\textsuperscript{\dag}}
\\
\\
 \textsuperscript{\dag}UNED Research Group in NLP and IR ({\tt nlp.uned.es}), Madrid, Spain
 \\
 \small{
   \textbf{Correspondence:} \href{gmarco@lsi.uned.es}{gmarco@lsi.uned.es}
 }
}

\begin{document}
\maketitle

\input{sections/0_abstract}

\input{sections/1_introduction}

\input{sections/2_related_work_finetuned}

\input{sections/3_problem_definition}

\input{sections/4_methodology}

\input{sections/6_results}

\input{sections/7_conclusions}

%\clearpage

\input{sections/X_0_limitations}

\input{sections/Z_acknowledgements}
\bibliography{custom}

\appendix
\input{sections/X_1_all_metrics}

\input{sections/X_3_correlation_analysis}
\input{sections/X_2_dataset_limitations}

\end{document}

%% file: sections/0_abstract.tex
\begin{abstract}
Recent studies comparing AI-generated and human-authored literary texts have produced conflicting results: some suggest AI already surpasses human quality, while others argue it still falls short. We start from the hypothesis that such divergences can be largely explained by genuine differences in how readers interpret and value literature, rather than by an intrinsic quality of the texts evaluated. Using five public datasets (1,471 stories, 101 annotators including critics, students, and lay readers), we (i) extract 17 reference-less textual features (e.g., coherence, emotional variance, average sentence length...); (ii) model individual reader preferences, deriving feature importance vectors that reflect their textual priorities; and (iii) analyze these vectors in a shared “preference space”. Reader vectors cluster into two profiles: \textit{surface-focused readers} (mainly non-experts), who prioritize readability and textual richness; and \textit{holistic readers} (mainly experts), who value thematic development, rhetorical variety, and sentiment dynamics. Our results quantitatively explain how measurements of literary quality are a function of how text features align with each reader’s preferences. These findings advocate for reader-sensitive evaluation frameworks in the field of creative text generation.\footnote{Code, data, and results: \url{https://github.com/grmarco/the-reader-is-the-metric}}
\end{abstract}

%% file: sections/1_introduction.tex
\section{Introduction}

\input{tables/1_llms_vs_writers}

Large Language Models (LLMs) are increasingly able to generate short stories that resemble human writing, leading to a growing number of evaluation studies based on reader judgments.  As Table~\ref{tab:comparison} shows, the results of such studies are mixed. Some find that lay readers often prefer AI-generated texts—even over canonical authors like Shakespeare \citep{porter_ai-generated_2024}—or rate a small language model above the average human writer in creative tasks \citep{marco_small_2025}. Others find that expert judges consistently favor human texts. Critics rank Patricio Pron’s stories above those written by GPT-4 \citep{marco_pron_2024}, human-written texts satisfy far more creativity criteria in the Torrance Test \citep{chakrabarty_art_2024}. \citet{gomez-rodriguez_confederacy_2023} find a mixed picture: top commercial LLMs match or even surpass humans in fluency and coherence, but humans keep the edge in creativity and nuanced humour. What drives these apparently opposing judgments, and can this variation be measured in a systematic way? Answering this question is the core motivation of our work.

These studies implicitly assume that aggregating ratings yields an objective measure of literary quality within an evaluative group. This assumes the existence of a shared standard, treating differences in judgment as noise.  We adopt a \emph{perspectivist} view \citep{plank_problem_2022,bizzoni_predicting_2022}: what counts as “good writing” depends on the reader, and this variation is meaningful.

%To study this variation systematically, we focus on one key layer of literary evaluation: the intrinsic properties of the texts themselves. 

While many factors can influence literary judgment, including references to other texts, political alignment, or familiarity with genre conventions, this study focuses on intrinsic textual features: measurable properties related to style, structure, and content. These features provide a reasonable starting point for modeling variation in reader preferences, as they can be extracted directly from the text and interpreted consistently across corpora. 

Inspired by multi-criteria decision-making theory \citep{triantaphyllou2000multi}, we conceptually frame each reader’s evaluation as an additive utility function over intrinsic textual features. For each reader, we estimate a weight vector that best predicts their observed ranking of texts. These reader-specific weights provide an interpretable representation of literary preferences, which can be clustered to reveal shared evaluative profiles.

This framework lets us examine when, how, and for whom AI stories become indistinguishable from human ones, and clarifies the gap between expert and non-expert opinions.

To guide our investigation into these aspects, this work seeks to answer several research questions: \textbf{(RQ1)} Do AI and human stories differ in measurable textual features across datasets? \textbf{(RQ2)} Can reader preferences be modeled from these features? \textbf{(RQ3)} Do experts and non-experts prioritize different features? And \textbf{(RQ4)} do these preference patterns remain consistent across datasets?

In addressing these research questions, this paper makes the following contributions:

\begin{itemize}[leftmargin=1em]
    %\item We align and normalize five public datasets of blind short story evaluations, enabling cross-dataset modeling of reader preferences across LLMs, text types, and annotator backgrounds.
    
    \item We compile and implement an open-source set of interpretable, reference-less textual metrics that capture stylistic, structural, and semantic properties relevant to literary judgment.
    
    \item We introduce a method to model reader preferences as weighted combinations of the above textual metrics, enabling interpretable analysis of audience-specific judgment.
    
    \item Using five public datasets, we show that AI-generated texts gain favor when their feature profiles align with reader-preferred textual characteristics, which often overlap with or emulate features found in valued human-authored stories. 
    %This indicates the potential for LLMs not only to mimic human writing but also to be strategically directed towards textual feature combinations known to be favored by particular reader groups. 
    
    \item We demonstrate that individual reader preferences, when modeled from textual features, cluster into distinct evaluative profiles that subsequently show a strong correlation with reader expertise. Specifically, emergent clusters reveal two primary reader types: one, largely composed of lay readers, prioritizes linguistic richness and textual accessibility (e.g., sentence complexity, readability, lexical diversity). The other, predominantly comprising experts, assigns greater weight to thematic development, expressive stylistics, and sentiment arcs. These emergent, divergent evaluative criteria are identified as a key factor in explaining the conflicting results observed in prior studies comparing AI and human writing.
\end{itemize}

%\subsection{Contributions}

%This paper contributes to the study of AI-generated literature in several ways:

%\begin{itemize}
 %   \item We standardize and align five existing datasets of blind short story evaluations, enabling consistent modeling across reader types, model sizes, and text types.
    
  %  \item We release a comprehensive set of interpretable, reference-less metrics to characterize literary texts across stylistic, structural, and semantic dimensions.
    
   % \item We propose a method for modeling individual reader preferences as feature-weighted functions, enabling fine-grained analysis of how different audiences interpret and value literary texts. 
    
    %\item We show that AI-generated texts become equally likely to be selected as reader favorites when their feature profiles align with those of human-authored stories—suggesting that LLMs can be steered into regions of the textual space typically associated with human writing.

    %\item We identify systematic differences between expert and non-expert readers: lay audiences tend to favor fluency and surface-level features, while experts prioritize structural and thematic dimensions.  These differences emerge naturally from the data as consistent clusters of evaluative profiles across datasets. This helps explain apparent contradictions in prior work, and reveals the limits of Turing-test-style setups that seek a universal benchmark for literary quality.
%\end{itemize}

%% file: tables/1_llms_vs_writers.tex
\begin{table*}
    \centering
    \footnotesize
    \renewcommand{\arraystretch}{1.1}
    \setlength{\tabcolsep}{4pt}
    \begin{tabular}{ l l l l c }
        \toprule
        \textbf{Paper} & \textbf{AI writer} & \textbf{Human Writers} & \textbf{Assessors} & \textbf{AI surpass Humans} \\
        \midrule
        \cite{marco_small_2025} & SLM & Average Writers & Average Readers &  \checkmark \\
        
        \cite{porter_ai-generated_2024} & LLM & Top Writers & Average Readers & \checkmark \\
        \cite{gomez-rodriguez_confederacy_2023} & LLM & Writing Students & Writing Students & $\approx$ \\
        \cite{chhun_language_2024} & LLM & Average Writers & LLM & $=$ \\

        \cite{chakrabarty_art_2024} & LLM & Professional Writers & Literary Experts & \textcolor{red}{$\times$} \\
        \cite{marco_pron_2024} & LLM & Professional Writers & Literary Experts & \textcolor{red}{$\times$} \\
       
        \hline

    \end{tabular}
    \caption{Comparison of AI-generated text performance against human writers.}
    \label{tab:comparison}
\end{table*}

%% file: sections/2_related_work_finetuned.tex
\section{Related Work}

Research at the intersection of Artificial Intelligence (AI) and literature has explored several key areas. A prominent line of work directly compares the perceived quality of AI-generated and human-authored texts, often framing the comparison as a competition \cite{gomez-rodriguez_confederacy_2023,porter_ai-generated_2024, chakrabarty_art_2024, marco_pron_2024,marco_small_2025}. These studies reveal a complex picture, with outcomes varying significantly depending on the type of text (e.g., poetry vs. short stories), the expertise of the evaluators (e.g., average readers vs. literary critics), and the specific AI models employed. For instance, some studies indicate that AI-generated texts are sometimes preferred by lay readers \cite{marco_small_2025,porter_ai-generated_2024}, while others, particularly those involving literary experts, consistently show a preference for human-authored texts \cite{marco_pron_2024,chakrabarty_art_2024}.

Building on the observation of varied outcomes in quality perception, recent research delves deeper by quantifying specific aspects like creativity. The "Creativity Index," for instance, assesses linguistic originality by how much a text can be reconstructed from web snippets, revealing that professional human authors exhibit a 66.2 \% higher index than LLMs \cite{lu-etal-2025-salieri}. Furthermore, the process of aligning LLMs with human preferences through methods like RLHF has been shown to inadvertently reduce their Creativity Index by about 30.1\% \cite{lu-etal-2025-salieri,yu-etal-2025-diverse}. This observation implies a trade-off: enhancing safety or coherence might occur at the expense of originality \cite{yu-etal-2025-diverse}, a dynamic particularly relevant when considering how different reader profiles, especially experts valuing originality, might assess these outputs.

The current work builds upon a foundation of research in computational literary studies that seeks to identify textual correlates of literary quality. Studies have explored how features such as global coherence, local unpredictability, sentiment arcs, and emotional volatility differentiate between canonical literature, bestsellers, and other categories \cite{bizzoni_fabulanet_2024, feldkamp_measuring_2024, bizzoni_global_2024, moreira_modeling_2023}. These insights inform the selection of our implemented textual metrics. However, our study does not directly use these metrics to predict literary quality. Instead, they are leveraged to model how individual readers differentially weight these properties, reflecting their unique quality model. 

%Inspiration is also drawn from general text generation evaluation, which considers features such as coherence and fluency \cite{celikyilmaz2020evaluation}.
The importance of perspectivism in literary evaluation has been emphasized in prior work. \citet{bizzoni_predicting_2022} propose a "mild perspectivist" approach that aggregates literary judgments within socially or professionally coherent reader classes. They show that each class yields internally consistent quality judgments and argue that such reader classes are the natural unit for modelling literary value. Our study extends this perspectivist view by modeling reader preferences at the individual level, not based on predefined categories, but inferred directly from annotated comparisons and textual features. This enables a finer-grained, data-driven reconstruction of evaluative profiles, which can then be clustered to reveal latent reader types without pre-established assumptions.

%The importance of interpretability in literary evaluation has been emphasized in prior work. \citet{bizzoni_predicting_2022} advocate for a perspectivist approach, arguing against both monolithic ground truths and fully idiosyncratic views by grouping readers into classes (e.g., by gender or professional background) that display internal consistency in quality judgments. Our study builds upon this perspective but moves further: rather than defining fixed reader groups a priori, we infer evaluative profiles directly from individual preferences, modeling and clustering them based on the textual features they prioritize. This allows us to explain reader variation quantitatively and uncover stable preference structures without relying on demographic labels.

%The importance of interpretability in literary evaluation has been emphasized in prior work. \citet{bizzoni_predicting_2022} advocate for a "perspectivist" approach, acknowledging that literary quality is contingent on reader background and expectations. Our study moves beyond simply recognizing the subjectivity of literary judgment to actively modeling and explaining such subjectivity.

%% file: sections/3_problem_definition.tex
\section{Problem Definition}
\label{sec:problemdef}

We consider a collection of $t$ texts $\mathcal{T} = \{x_1, x_2, \dots, x_t\}$, each described by a $d$-dimensional vector of intrinsic, reference-less features
\[
\mathbf{x}_i = (x_{i1}, x_{i2}, \dots, x_{id}) \in \mathbb{R}^d,
\]
capturing several dimensions related to coherence, originality, readability, sentiment arcs, and stylistic complexity (detailed in Section~\ref{sec:metrics}).

Additionally, we have a set $\mathcal{L} = \{r_1, r_2, \dots, r_l\}$ of $l$ readers. Each reader $r_j\!\in\!\mathcal{L}$ provides evaluations for texts $x_i$. These diverse raw evaluations (e.g., multi-dimensional Likert scores, collections of binary judgments) are first aggregated into a single raw preference score $S_j(x_i)$ for each text $x_i$ evaluated by reader $r_j$. To enable comparison across heterogeneous scales and readers, these raw scores $S_j(x_i)$ are then mapped to a continuous \emph{preference value} $\rho_j(x_i) \in [0,1]$ via per-reader min-max normalisation:
\begin{equation}
    \rho_j(x_i) \;=\;
\frac{S_j(x_i) - S^{\min}_j}{S^{\max}_j - S^{\min}_j},
\label{eq:rho_definition_in_problemdef}
\end{equation}
where $S^{\min}_j$ and $S^{\max}_j$ are the minimum and maximum raw preference scores $S_j(\cdot)$ assigned by $r_j$ within that dataset. Sorting texts by $\rho_j(x_i)$ yields a reader-specific ranking $\pi_j$.

\paragraph{Reader Preference Centroid}
For each reader $r_j$, their reader preference centroid or ``paradigmatic text'' is profiled by $\mathbf{x}_j^*$, the $\rho_j(x_i)$-weighted average of features from their top-rated texts $X_j^{top}$ (e.g., top 25\% of their evaluations):
\begin{equation}
\label{eq:paradigmatic}
    \mathbf{x}_j^{*} \;=\; \frac{\sum_{x_i \in X_j^{top}}\,\rho_j(x_i)\,\mathbf{x}_i}{\sum_{x_i \in X_j^{top}}\,\rho_j(x_i) + \epsilon}.
\end{equation}
These profiles capture preferred textual characteristics within a specific dataset.

\paragraph{Reader Utility Model} Inspired by multi-criteria decision-making theory
\citep{triantaphyllou2000multi},
we posit that each reader implicitly combines textual features through an
\emph{additive utility function}:
\begin{equation}
\label{eq:utility}
U_j(x_i) \;=\; \sum_{f=1}^{d} w_{jf}\,x_{if},
\end{equation}
where $\mathbf{w}_j = (w_{j1}, \dots, w_{jd}) \in \mathbb{R}^d_{\ge 0}$
encodes the conceptual importance reader $r_j$ assigns to feature~$f$.
A higher $U_j(x_i)$ indicates better alignment with $r_j$’s preferences. Eq.~\ref{eq:utility} serves as a conceptual starting point; practical estimation could implement non-linear models (as we do in Section~\ref{sec:methodology}) to derive empirical proxies for $\mathbf{w}_j$ as feature importances.

\bigskip
This formalisation sets the stage for our empirical study: (1) characterizing human- and AI-authored texts within their shared feature space to examine their properties and interrelations, and (2) inferring reader-specific preferences (via proxies for $\mathbf{w}_j$ and profiles $\mathbf{x}_j^*$) to analyse patterns of literary preference. 

%% file: sections/4_methodology.tex
\section{Methodology}
\label{sec:methodology}
Our methodology proceeds by: (1) analyzing the datasets and extracting textual features; (2) processing reader evaluations into unified preference scores and pairwise data; (3) conducting an exploratory analysis of text features and initial reader preferences to predict preferences (RQ1); and (4) training individualized Random Forest models (RQ2), from which feature importances are extracted and clustered to identify reader profiles (RQ3-4). These steps are detailed below.

\subsection{Data: Corpora \& Annotation Design}
\label{sec:dataset_selection}

The five datasets selected for this study offer complementary perspectives on human evaluation of AI-generated literary texts. Each provides explicit reader-level judgments, includes both AI- and human-authored short narratives, and retains the full textual content for feature extraction. Table \ref{tab:datasets} summarizes the main characteristics of each one.

\textsc{slm} \cite{marco_small_2025} comprises 122 movie synopses written by either humans or a fine-tuned BART-large model \cite{lewis-etal-2020-bart}, evaluated by 68 lay readers across five dimensions: creativity, readability, relevance, understandability, and attractiveness. It compares the creative output of small language models with that of average human writers under varying disclosure conditions regarding the author’s.

\textsc{hanna} \cite{chhun_human_2022} contains over 1,000 narrative texts, each evaluated by ten Mechanical Turk workers along multiple 5-point Likert scales—such as fluency, coherence, and overall quality.

\textsc{confederacy} \cite{gomez-rodriguez_confederacy_2023} 65 short stories (250-1200 words) from a single prompt, 5 human-authored, 60 AI-generated (5 per model from 12 LLMs). Each rated by 2 independent raters on 10 dimensions (1300 ratings total).

\textsc{PronvsPrompt} \cite{marco_pron_2024} comprises 180 fictional film synopses of approximately 600 words. Sixty were authored by Spanish writer Patricio Pron, and 120 were generated by GPT-4. Six literary experts—editors, scholars, and instructors—evaluated 120 stories each, blind to authorship. Their assessments followed a rubric grounded in Margaret Boden’s theory of creativity, rating dimensions such as originality, narrative voice, and attractiveness.

\textsc{ttcw} \cite{chakrabarty_art_2024} consists of 48 short stories derived from 12 one-sentence plot prompts. For each prompt, one human-authored story from \textit{The New Yorker} and three AI-generated stories (by GPT-3.5, GPT-4, and Claude 1.3) were produced, yielding 36 machine texts and 12 human texts. Each story was evaluated by three experts—creative writing professors, literary agents, and MFA-trained authors—using a 14-item binary rubric adapted from the Torrance Test of Creative Thinking (TTCT), covering fluency, flexibility, originality, and elaboration.

\input{tables/5_dataset_descriptions}

\subsection{Feature Extraction}
\label{sec:metrics}

To characterize the short stories, we selected a set of reference-less metrics grounded in prior work on literary text evaluation and stylometric analysis \cite{celikyilmaz2020evaluation, bizzoni_fractality_2022, feldkamp_measuring_2024}. Rather than prescriptive quality indicators, these metrics are descriptive features designed to capture salient textual properties. We selected them based on three criteria: (1) theoretical relevance to literary characterization, (2) interpretability, and (3) low redundancy (Pearson $|r| < 0.7$, see Appendix~\ref{appx:correlation}), which ensures a diverse yet concise feature set and mitigates multicollinearity. The selected metrics, formally defined in Appendix~\ref{app:metrics}, encompass:

\begin{itemize}[leftmargin=1em]
    \item \textbf{Linguistic fluency and readability}: including \textit{average sentence length} as a basic measure of syntactic load; the \textit{SMOG index} \citep{mclaughlin1969smog} for syllable-based readability assessment; and \textit{MTLD} \citep{mccarthy2010mtld} for a measure of lexical diversity.

    \item \textbf{Structural and thematic coherence}: such as \textit{entity coherence}, which models narrative continuity based on entity-grid transitions \citep{barzilay2008entity}; \textit{local coherence embeddings}, assessing semantic linkage between adjacent sentences using their distributed representations \citep{li2014coherence}; as well as \textit{thematic graph density} and \textit{thematic entropy}, which quantify the connectivity \citep{newman2010networks} and diversity \citep{shannon1948communication} of topical structures discovered via Latent Dirichlet Allocation \citep{blei2003lda, roder2015topic}.

    \item \textbf{Sentiment dynamics}: including \textit{average sentiment}, \textit{sentiment variance}, and \textit{emotional volatility}. These reflect the overall tonal polarity and its consistency, alongside the variability of dominant emotions across the narrative arc, as identified by fine-tuned RoBERTa-based models \citep{hartmann2023more, demszky2020goemotions, liu2019roberta}.

    \item \textbf{Originality and Predictability}: approximated by the \textit{log-likelihood} of the text under a large language model (\texttt{meta-llama/Llama-3.2-3B-Instruct}; \citep{aiatmeta2024llama3models}). Lower predictability is taken as an indicator of greater textual originality.
\end{itemize}

While the feature set does not cover higher-order literary constructs—such as narrative framing, ideological alignment, or intertextual reference—it provides a simple, transparent and replicable foundation for approximating reader preferences across diverse corpora.

\subsection{Deriving Reader-wise Preference Data}
\label{subsec:deriving_preference_data}

As outlined in Section~\ref{sec:problemdef}, raw reader evaluations $S_j(x_i)$ are first processed into an aggregated form. The specific aggregation method for $S_j(x_i)$ varies by dataset:
\begin{itemize}[leftmargin=1.5em, itemsep=0pt, topsep=3pt, partopsep=0pt]
    \item For datasets with multi-dimensional Likert-scale evaluations (\textsc{slm}, \textsc{hanna}, \textsc{confederacy}, \textsc{pronvsprompt}), $S_j(x_i)$ is the sum of scores across all evaluated dimensions for text $x_i$ by reader $r_j$.
    \item For datasets with binary verdicts on $N_{criteria}$ criteria (e.g., \textsc{ttcw}), $S_j(x_i)$ is the count of positive (``Yes'') verdicts.
\end{itemize}
These aggregated raw scores $S_j(x_i)$ are then min-max normalized per reader to yield continuous preference values $\rho_j(x_i) \in [0,1]$ using Eq.~\ref{eq:rho_definition_in_problemdef}. These $\rho_j(x_i)$ values are subsequently used to compute reader preference centroids (Section~\ref{subsec:exploratory_paradigmatic}) and serve as the basis for generating pairwise preference data used in training our individualized learning-to-rank models, a process detailed in Section~\ref{subsec:learning_preference_models}. % Referencia a la siguiente sección

\subsection{Exploratory Analysis and Reader Preference Centroids (RQ1)}
\label{subsec:exploratory_paradigmatic}

To investigate RQ1 concerning systematic differences between AI-generated and human-authored texts and their relation to initial reader preferences, we first compute each reader $r_j$'s reader preference centroid $\mathbf{x}_j^*$. This vector, calculated as per its formula (Eq.~\ref{eq:paradigmatic}) using the derived preference values $\rho_j(x_i)$ (Section~\ref{subsec:deriving_preference_data}), represents $r_j$'s Reader Preference Centroid. Crucially, these $\mathbf{x}_j^*$ profiles are defined within the $d$-dimensional text feature space of a specific dataset, reflecting $r_j$'s high-preference textual characteristics within that particular corpus. 
%While insightful for understanding which concrete textual features are valued in a given context, these dataset-specific profiles do not directly allow for comparison of underlying preference \textit{structures} across different corpora; this motivates the subsequent learning of individualized preference models (Section~\ref{subsec:learning_preference_models}).

Subsequently, we employ Principal Component Analysis (PCA) for exploratory visualization. We project all text feature vectors $\mathbf{x}_i$ onto a common low-dimensional space (typically $\mathbb{R}^2$) to examine AI vs. human text distributions. PCA loadings help identify key differentiating features. Onto this same space, we project the dataset-specific reader preference centroid $\mathbf{x}_j^*$ to observe how these reader-preferred text characteristics align with the global text distributions. This combined visualization provides initial insights for RQ1 (detailed in Section~\ref{sec:results_rq1}).

\subsection{Learning Individualized Preference Models (RQ2)}
\label{subsec:learning_preference_models}

While Text Reader Centroids are insightful for understanding which specific textual features are valued in a given context, these dataset-specific profiles do not directly allow for comparison of underlying preference \textit{structures} across different corpora (RQs2-4); this motivates the subsequent learning of individualized preference models. The primary challenge in learning these preference models lies in the nature of the available data: many readers have evaluated only a limited number of texts (see Table~\ref{tab:datasets}), making direct ranking functions difficult to train robustly due to sparse per-reader evaluations. Thus, we use a supervised pairwise learning-to-rank approach, a common strategy that converts ranking to binary classification.

To generate the training data for this classification task, we use the normalized preference values $\rho_j(x_i)$ derived in Section~\ref{subsec:deriving_preference_data}. Specifically, for each reader $r_j$ and any two distinct texts $x_a, x_b$ they evaluated where $\rho_j(x_a) \neq \rho_j(x_b)$, we generate balanced training instances: $(\mathbf{x}_a - \mathbf{x}_b, 1)$ if $\rho_j(x_a) > \rho_j(x_b)$ (and its opposite $(\mathbf{x}_b - \mathbf{x}_a, 0)$), or vice-versa. Pairs with equal preference scores ($\rho_j(x_a) = \rho_j(x_b)$) are discarded.

For each such pair, we derive the feature difference $\mathbf{x}_a - \mathbf{x}_b \in \mathbb{R}^d$. We then train a separate Random Forest classifier \citep{pedregosa2011scikit, breiman2001random} for each reader $r_j$ using these difference vectors and their corresponding binary preference labels. Initially, we tested Logistic Regression and Random Forest. Random Forest outperformed Logistic Regression, achieving higher F1 scores on the test set (Appendix~\ref{appx:correlation}). Likely, literary judgment does not emerge from simple linear relationships between textual features but rather from non-linear relationships and complex feature interactions, which Random Forest better approximates.

\subsection{Interpreting Preferences and Identifying Reader Profiles (RQ3-4)}
\label{subsec:preference_profiles}

Once a Random Forest model is trained for each reader \( r_j \), we extract feature importance values to determine which attributes were most influential in distinguishing \( \mathbf{x}_i \) from \( \mathbf{x}_{i'} \). Specifically, we use the mean decrease in impurity (MDI) which measures how much each feature reduces the weighted impurity (Gini entropy) across all trees in the forest \citep{breiman2001random}. These importance values serve as empirical proxies for the conceptual weights \( w_{jf} \) in the additive utility model discussed in Section \ref{sec:problemdef}, providing an interpretable, albeit approximated, representation of the reader’s preference structure. Clustering these importance vectors across many readers helps reveal broader patterns. For instance, different subsets of readers (e.g., literary experts vs.\ non-experts) may prioritize distinct sets of textual attributes, reflecting variations in evaluative criteria and reading strategies.

In summary, our methodology integrates an exploratory stage, where PCA is used to examine textual feature distributions, and a second stage in which we train individualized pairwise preference models to learn generalizable evaluative patterns and quantify the importance of textual features for each reader across datasets.

%% file: tables/5_dataset_descriptions.tex
\begin{table*}[t]
\centering
\small
\setlength{\tabcolsep}{5pt}
\begin{tabular}{lcrrrll}
\toprule
\textbf{Alias} &
\textbf{\# Texts} &
\textbf{AI/H} &
\textbf{\# Readers} &
\textbf{Scale} &
\textbf{Annotator BG} &
\textbf{Source} \\
\midrule
\textsc{slm}            & 122 & 61/61  & 68 & 7-pt Likert & Lay (biz students) & \citet{marco_small_2025} \\
\textsc{hanna}          & 1 056 & 960/96 & 10 & 5-pt Likert & Lay (MTurk)        & \citet{chhun_human_2022} \\
\textsc{confederacy}    & 65  & 60/5   & 10 & 10-pt Likert & Lit. students      & \citet{gomez-rodriguez_confederacy_2023} \\
\textsc{ttcw}           & 48  & 36/12  & 7  & 14-item bin. & Critics / writers  & \citet{chakrabarty_art_2024} \\
\textsc{pronvsprompt}   & 180 & 120/60 & 6  & 7-pt Likert  & Critics            & \citet{marco_pron_2024} \\
\bottomrule
\end{tabular}
\caption{Corpora used to model reader preferences. “AI/H” is the split
between AI-generated and human-authored texts. Full corpus statistics,
including total pairwise comparisons and AI-preference rates, are in
Appendix~C.}
\label{tab:datasets}
\end{table*}

%% file: sections/6_results.tex
\section{Results and Discussion}

In this section, we examine whether AI- and human-authored texts differ measurably in their feature distributions, how these distributions relate to reader preferences, and whether reader-specific models can capture consistent patterns of evaluation. We structure the analysis around our four research questions.

\begin{figure*}[t]
  \captionsetup[sub]{margin=0pt,skip=2pt}
  \centering
  
  % ---------- Fila 1 ----------
  \makebox[\textwidth][c]{%
    \begin{subfigure}[t]{0.34\textwidth}
      \includegraphics[width=\linewidth]{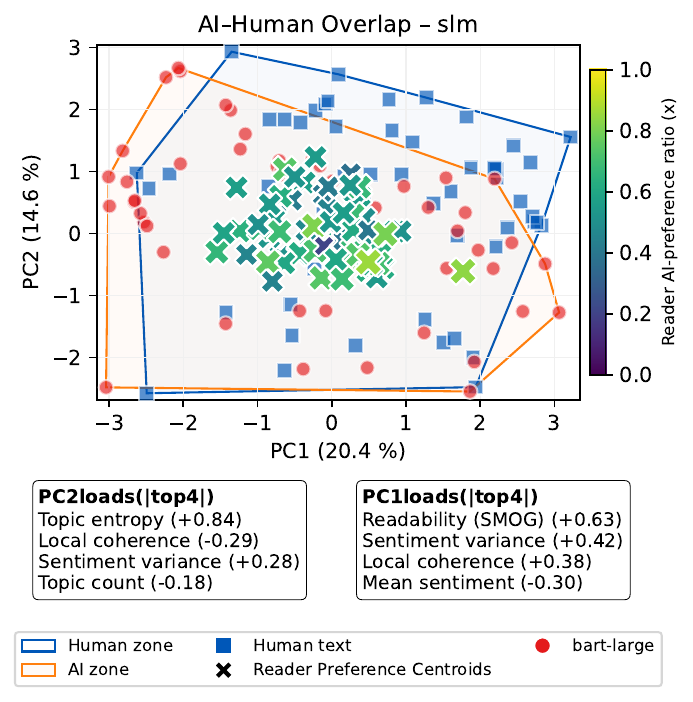}
      \caption{\small \textsc{SLM}\\\textbf{AI preference: 57.59\%}}
      \label{fig:slm}
    \end{subfigure}\hspace{6em}
    \begin{subfigure}[t]{0.34\textwidth}
      \includegraphics[width=\linewidth]{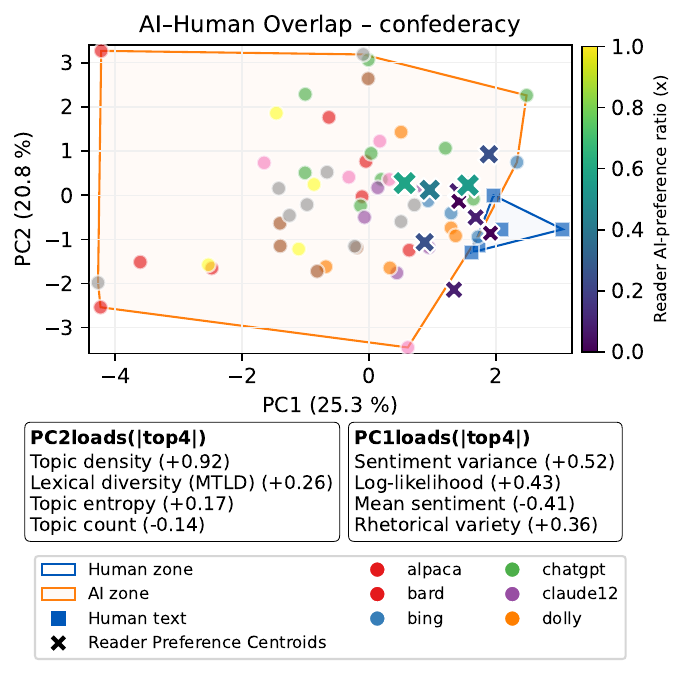}
      \caption{\small Confederacy\\\textbf{AI preference: 22.95\%}}
      \label{fig:confederacy}
    \end{subfigure}%
  }

  \vspace{0.8em} % espacio vertical opcional

  % ---------- Fila 2 ----------
  \makebox[\textwidth][c]{%
    \begin{subfigure}[t]{0.34\textwidth}
      \includegraphics[width=\linewidth]{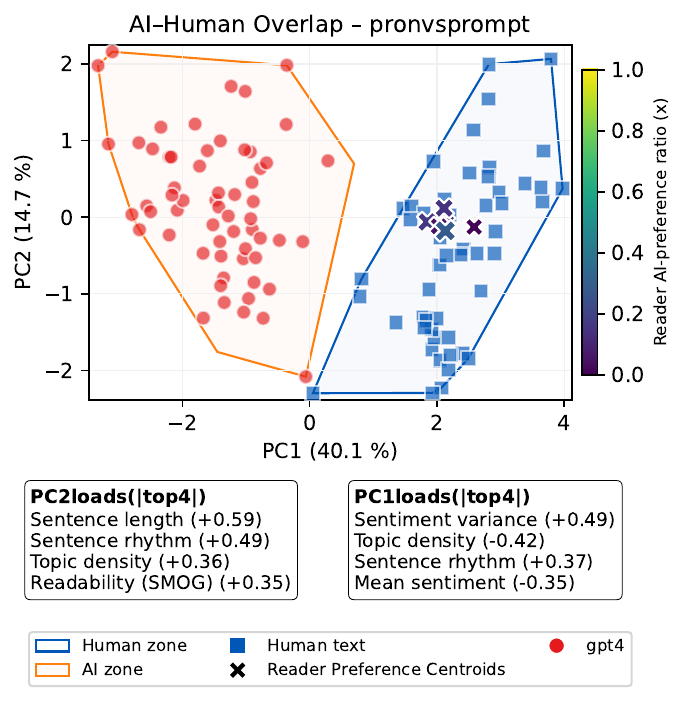}
      \caption{\small \textsc{PronvsPrompt}\\\textbf{AI preference: 13.12\%}}
      \label{fig:pronvsprompt}
    \end{subfigure}\hspace{6em}
    \begin{subfigure}[t]{0.34\textwidth}
      \includegraphics[width=\linewidth]{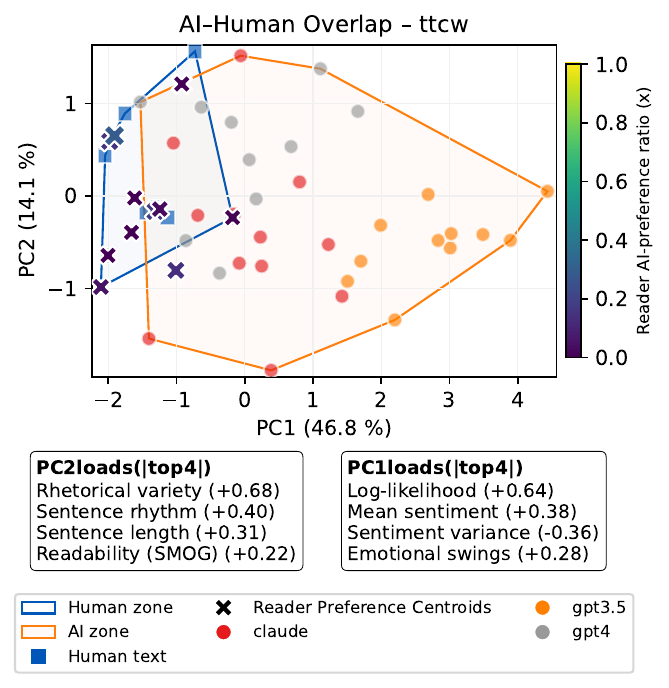}
      \caption{\small \textsc{ttcw}\\\textbf{AI preference: 2.16\%}}
      \label{fig:ttcw}
    \end{subfigure}%
  }
\caption{ PCA projection of textual feature vectors for AI-generated and human-authored stories across four datasets (a-d), sorted by decreasing AI preference rate. Each point is a story. Axes (PC1, PC2) are principal components; adjacent boxes detail their top feature loadings. Convex hulls show 'AI' (orange) and 'Human' (blue/darker outline) textual feature zones, indicating overlap or separation. 'Reader Preference Centroids' (X marks) represent the average features of highly-rated stories for each reader; their size and color intensity correspond to the individual reader's AI preference ratio. The overall 'AI preference' percentage shows how often readers favored AI texts within that dataset. The figure illustrates how text separability and centroid alignment with these zones correlate with these AI preference rates across datasets.}
  %\caption{PCA projection of textual feature vectors for AI-generated and human-authored stories across four datasets (a-d), sorted by decreasing AI preference rate. Each point represents a single story. The axes (PC1, PC2) represent the two principal components capturing the most variance in textual features; the boxes detail the top features loading onto each component, indicating what textual characteristics primarily define that dimension. Convex hulls delineate the 'AI zone' (orange) and 'Human zone' (blue/darker outline), visualizing the typical feature space occupied by each author type and their degree of overlap or separation. 'Reader Preference Centroids' (X marks) indicate the average textual characteristics of stories highly-rated by readers within each dataset. The 'AI preference' percentage reflects how often readers favored AI-generated texts. The figure illustrates how the separability of AI and human texts in the feature space, and the alignment of reader preferences (centroids) with these textual regions, correspond to the observed AI preference rates across different datasets and AI models.}
  \label{fig:pca_maps_compact}
\end{figure*}

\subsection{Textual Separability, Feature Space Topography, and AI Preference Ratios}
\label{sec:results_rq1}

To assess differences between AI and human stories in measurable textual features and their relation to reader preferences (RQ1), we projected intrinsic story features from five datasets onto their first two principal components (PCA) (Figure~\ref{fig:pca_maps_compact}). These visualizations delineate 'AI zones' and 'Human zones' using convex hulls, with 'Reader Preference Centroids' (crosses) indicating aggregate preference tendencies for each reader. 

In the \textsc{slm} dataset (Figure~\ref{fig:slm}), PC1 (20.4\% variance) is driven by \textit{Readability (SMOG)} (+0.63) and \textit{Sentiment variance} (+0.42), while PC2 (14.6\% variance) is primarily influenced by \textit{Topic entropy} (+0.84). An extensive feature space overlap is observed between human-authored texts and those from the \texttt{bart-large} model. Reader Preference Centroids (crosses) are notably concentrated within this central overlapping region. A visual inspection suggests that the \texttt{bart-large} model generated more texts that fall into this main area of overlap. While approximately 22 of the 61 human texts appear distributed outside this primary overlapping zone, with many populating the distinct upper-right quadrant (characterized by lower local coherence and higher topic entropy, as indicated by PC2 loads), only around 11 of the 61 AI-generated texts seem similarly dispersed. Consequently, AI effectively "placed" more texts (approximately 50 out of 61) within this central overlapping feature space, compared to human authors (approximately 39 out of 61). Given that reader preferences are densely clustered within this same overlapping area, the AI's greater presence in this zone—where texts may be perceived as indistinguishable or even preferable according to the metrics—likely explains its higher preference rate of 57.59\%.
%In the \textsc{slm} dataset (Figure~\ref{fig:slm}), PC1 (20.4\% variance) is driven by \textit{Readability (SMOG)} (+0.63) and \textit{Sentiment variance} (+0.42), while PC2 (14.6\% variance) is primarily influenced by \textit{Topic entropy} (+0.84). Extensive feature space overlap is observed between human-authored texts and those from the \texttt{bart-large} model. While a small region in the upper-right quadrant (lower local coherence, higher topic entropy) is predominantly occupied by human texts, no clear general separation is evident. Reader Preference Centroids are concentrated in the larger shared region, indicating preferences are not strongly tied to unique authorial characteristics. This significant overall indistinguishability correlates with a high AI preference rate (57.59\%), suggesting general readers found the AI's aligned features comparable or even preferable.

The \textsc{confederacy} dataset (Figure~\ref{fig:confederacy}) presents human texts forming a cohesive cluster in the positive PC1 region (characterized by high \textit{Sentiment variance} (+0.52), \textit{Log-likelihood} (+0.43), and \textit{Rhetorical variety} (+0.36)) and the mid-to-low PC2 region (moderate \textit{Topic density} (+0.92)). AI-generated texts are widely dispersed, resulting in minimal overlap between the 'Human zone' and the majority of the 'AI zone'. This clear separation, unlike in \textsc{SLM}, correlates with a lower AI preference (22.95\%). The Reader Preference Centroids (crosses) indicate that while many preferences align with the human cluster, others are associated with specific AI texts. These preferred AI texts, though situated outside the primary human cluster, effectively emulate the strong positive PC1 characteristics (high sentiment variance and rhetorical variety) typical of the human-authored stories. Thus, in \textsc{confederacy}, AI preference appears driven not by general indistinguishability, but by the capacity of certain AI models to replicate specific, valued stylistic dimensions captured by PC1, which align with the evaluative criteria of the literature students.

\textsc{Pronvsprompt} (Figure~\ref{fig:pronvsprompt}) shows pronounced separation. Human texts (Pron) cluster in the high positive PC1 region, characterized by high \textit{Sentiment variance} (+0.49), marked \textit{Sentence rhythm} (+0.37), lower \textit{Topic density} (PC1 loading -0.42), and a less positive \textit{Mean sentiment} (PC1 loading -0.35). Conversely, GPT-4 texts occupy the negative PC1 region. With 'Human' and 'AI' zones being almost entirely disjoint, and Reader Preference Centroids exclusively within the 'Human zone', the literary critics' strong alignment with Pron's distinct features results in a very low AI preference (13.12\%).

In \textsc{ttcw} (Figure~\ref{fig:ttcw}), human texts form a distinct, compact cluster in the extreme negative PC1 region, signifying high LLM-based originality (\textit{Log-likelihood} loading for PC1 is +0.64), a more neutral/negative \textit{Mean sentiment} (PC1: +0.38), and high \textit{Sentiment variance} (PC1: -0.36). These texts also show high PC2 values (\textit{Rhetorical variety} (+0.68)). While most AI texts are in the positive PC1 space (more predictable, positive sentiment), a few Claude and GPT-4 instances fall within or near the 'Human zone', consistent with \citet{chakrabarty_art_2024} (Table \ref{tab:dataset_ttcw}). Despite this partial intrusion, Reader Preference Centroids remain almost exclusively with human texts, leading to a minimal AI preference (2.16\%) by expert evaluators.

In summary, this cross-dataset analysis (RQ1) reveals a nuanced landscape regarding the differences between AI-generated and human-authored stories and their relation to reader preferences. Measurable differences in textual features do exist, but their nature and magnitude vary considerably depending on the specific AI models, human comparison texts, and evaluative contexts. Literary appeal emerges from a dynamic interplay between a text's measurable characteristics, the degree to which AI can replicate or align with human-valued features, and the specific evaluative lens of the reader.

\subsection{Modeling Reader Preferences and Preference Clusters (RQ2)}
\label{ssec:reader_preference_modeling}

\begin{figure}%[ht]
    \centering

    % Subfigura 1
    \begin{subfigure}[b]{0.38\textwidth}
        \centering
        \includegraphics[width=\textwidth]{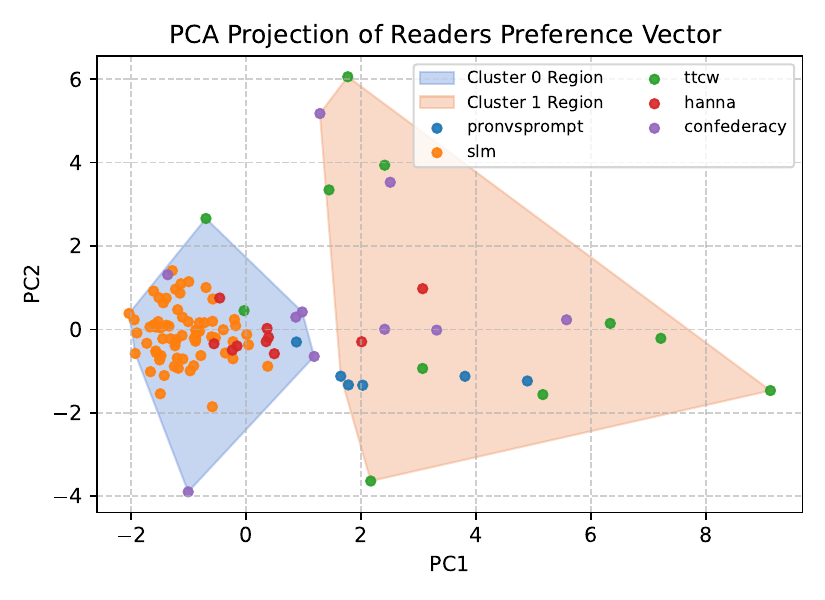}
        \caption{The PCA projection of how readers cluster according to their preference vectors (k-means, with k = 2)}
        \label{fig:pca_readers}
    \end{subfigure}
    \hfill
    % Subfigura 2
    \begin{subfigure}[b]{0.25\textwidth}
        \centering
        \includegraphics[width=\textwidth]{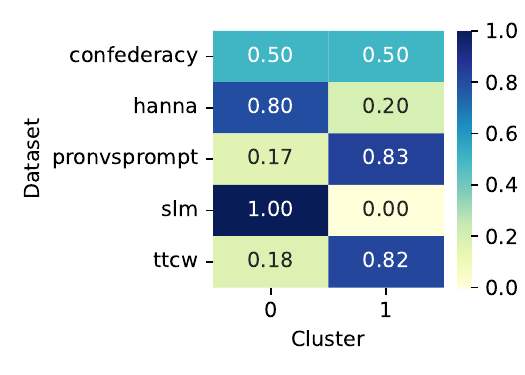}
        \caption{Distribution of readers across clusters by dataset.}
        \label{fig:cluster_membership_readers}
    \end{subfigure}

    \caption{Clustering of users based on feature importance vectors. \textsc{ttcw} and \textsc{PronvsPrompt}, evaluated by literary critics, predominantly fall into Cluster 1. \textsc{Hanna} and \textsc{SLM}, both composed of lay readers, are mainly grouped in Cluster 0. \textsc{Confederacy}, which includes writing students, shows a more balanced distribution across clusters, reflecting intermediate evaluation patterns.} %}
    \label{fig:clustering}
\end{figure}

To assess whether reader preferences can be modeled as weighted combinations of textual features, we first trained individual classifiers for each reader, as detailed in Section~\ref{subsec:preference_profiles}. For each reader, the trained classifier yields a feature importance vector, quantifying the relative weight each textual metric carries in predicting that reader's preferences. These individual feature importance vectors were then projected using PCA (Figure~\ref{fig:pca_readers}) to visualize the overall landscape of reader preference structures and subsequently clustered via k-means ($k=2$). The number of clusters was determined using standard validation techniques (see Section~\ref{subsec:preference_profiles}).

The resulting clusters exhibit a clear association with reader background, as shown by the distribution of readers per dataset across these clusters (Figure~\ref{fig:cluster_membership_readers}). Readers from the \textsc{SLM} and \textsc{Hanna} datasets (both comprising general audiences) are predominantly grouped in Cluster 0. In contrast, readers from \textsc{ttcw} and \textsc{PronvsPrompt} (primarily literary critics and experts) largely fall into Cluster 1. The \textsc{Confederacy} dataset, which includes literature students, shows a more mixed distribution across the two clusters, indicative of intermediate evaluation patterns.

These findings suggest that reader preferences reflect systematic, modelable tendencies that correlate with reader expertise. Crucially, these distinct preference clusters emerged solely from the modeling of evaluative judgments, without recourse to explicit reader metadata (such as self-declared expertise). 
%This indicates that the \textit{way} readers evaluate texts itself encodes discernible signals related to their background or expertise.

\subsection{Feature Prioritization by Reader Type (RQ3)}
\label{ssec:feature_prioritization}

\begin{figure}
    \centering
    \includegraphics[width=.45\textwidth]{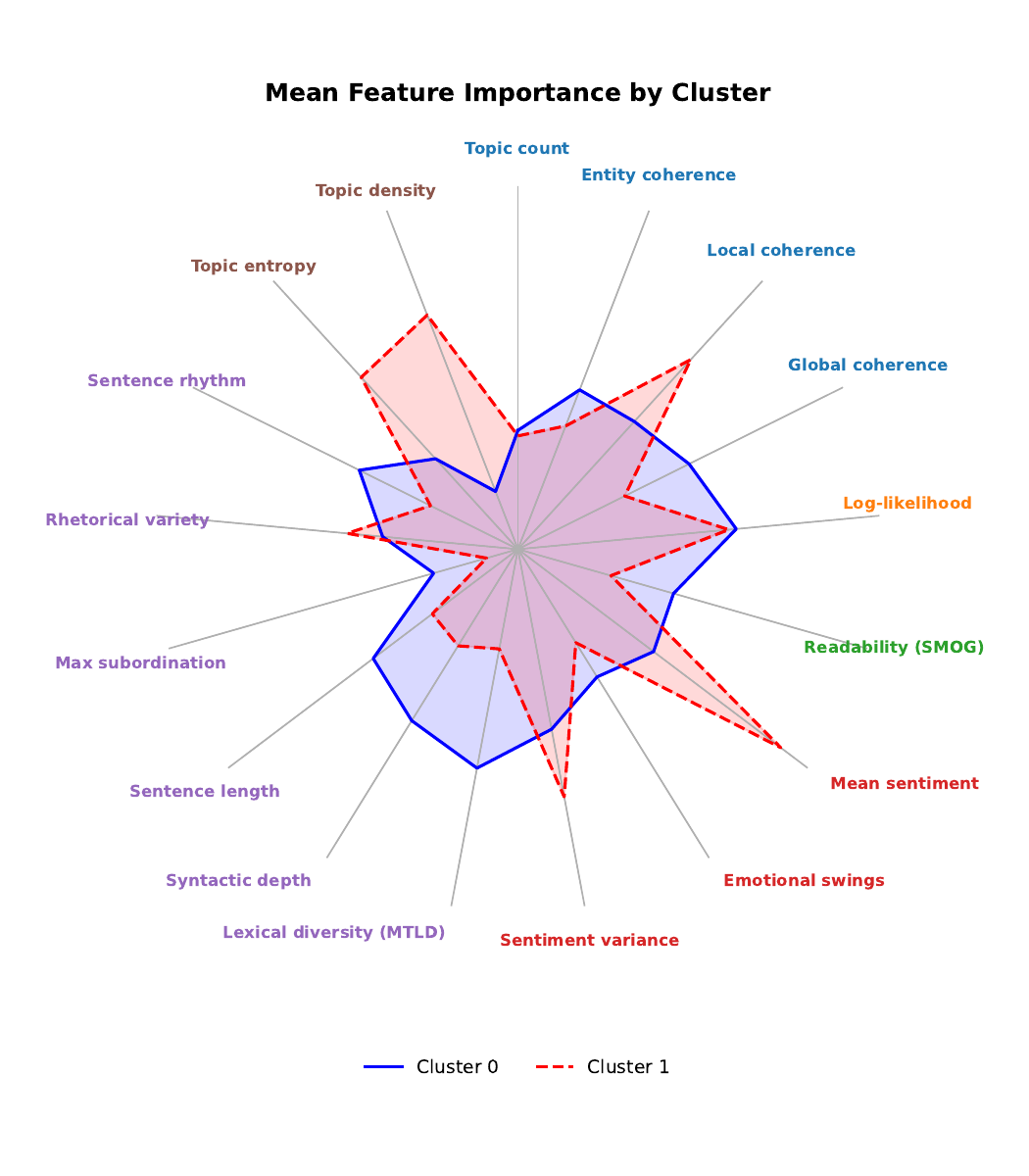}
    \caption{Radar chart illustrating the mean Random Forest feature importances for the two reader clusters. Higher values indicate greater importance assigned to a given textual feature when predicting preferences.}
    \label{fig:radar_importances}
\end{figure}

Figure~\ref{fig:radar_importances} visually contrasts the mean feature importance profiles for the two identified reader clusters. To understand the \textit{direction} of these preferences, we cross-reference these importance scores with the mean feature values from Table~\ref{tab:dataset_means} for texts typically favored by each reader type.

Cluster 0 (solid blue profile in Figure~\ref{fig:radar_importances}), mainly composed of lay readers (\textsc{SLM}, \textsc{Hanna}), emphasizes features related to linguistic complexity and textual accessibility. This group assigns notable importance to stylistic metrics such as \textit{Max subordination depth}, \textit{Sentence length}, \textit{Syntactic depth}, and \textit{Lexical diversity (MTLD)}. Analysis of the \textsc{SLM} dataset, where AI texts were often preferred by this cluster, shows these AI texts possessing, on average, higher maximum subordination, longer sentences, greater syntactic depth, and higher (less negative) MTLD scores (Table~\ref{tab:dataset_means}). The concurrent importance of \textit{Readability (SMOG index)}, for which these preferred AI texts exhibit lower scores (indicating easier readability), suggests a valuation for elaborate and lexically rich prose presented within an accessible framework.

In contrast, Cluster 1 (red dashed profile in Figure~\ref{fig:radar_importances}), predominantly including literary critics and professional writers (\textsc{ttcw}, \textsc{Pronvsprompt}), prioritizes features capturing thematic richness, expressive stylistics, sentiment dynamics, and coherence. High importance is attributed to \textit{Topic entropy}, \textit{Sentence rhythm}, and \textit{Rhetorical variety}. Human-authored texts, strongly favored by this cluster (Table~\ref{tab:dataset_means}), typically demonstrate higher thematic entropy and often more varied sentence rhythms. The importance of \textit{Rhetorical variety} is high, though human texts are not always highest on this specific metric compared to all AI. Furthermore, the entire suite of sentiment features (\textit{Mean sentiment}, \textit{Sentiment variance}) is highly prioritized. Human texts preferred by this group exhibit more neutral or less overtly positive mean sentiment, and higher sentiment variance. \textit{Local coherence} is also important; interestingly, human texts favored by experts show lower local coherence embedding scores, potentially reflecting an appreciation for more complex or less predictable local narrative transitions.

These divergent prioritizations, interpreted directionally using Table~\ref{tab:dataset_means}, directly address RQ3. Expert and non-expert readers weigh different textual dimensions and also exhibit preferences for specific feature levels. Non-experts (Cluster 0) favor texts with rich linguistic structure and good readability. Experts (Cluster 1) focus more on thematic development, expressive and varied stylistics, nuanced affective portrayal, and complex coherence patterns. The relatively similar importance assigned to \textit{Log-likelihood} (originality) by both clusters in Figure~\ref{fig:radar_importances} suggests it may be a generally valued characteristic. These distinctions underscore stable, interpretable differences in reader expectations.

\textbf{Cross-Dataset Consistency (RQ4).} The emergence of consistent textual distinctions (RQ1) and reader preference profiles (RQ2, RQ3) across the five diverse datasets underscores the stability and generalizability of our core findings. Crucially, the systematic clustering of reader preferences and their correlation with annotator expertise (RQ2) manifest despite the heterogeneity in prompts, AI models, and specific annotator groups across datasets.

%% file: sections/7_conclusions.tex
\section{Conclusions and Future Work}
\label{sec:conclusions_future_work}
This study demonstrates that literary quality judgments emerge from a structured interplay between a text's features and individual reader perception. We reveal two key drivers of literary preference: first, AI-generated stories gain favor when their textual profiles successfully emulate characteristics valued by specific reader groups, often aligning with features present in human-authored narratives that resonate with those readers; second, distinct reader communities consistently prioritize different textual dimensions. The preference for AI-generated texts, therefore, often hinges on their ability to exhibit the feature combinations favored by a particular reader profile. 

These stable, divergent preference profiles across datasets suggest that the notion of a universal standard for "good writing" may be an oversimplification.  Instead, our interpretable framework offers a path toward more nuanced evaluations of LLM-generated texts.

%% file: sections/X_0_limitations.tex
\section{Scope and Limitations}

This study investigates how literary preferences toward AI- and human-authored stories emerge from the interaction between intrinsic textual features and individual reader priorities. While our approach enables a systematic, interpretable, and reproducible analysis, it is not without limitations.

First, the analysis is restricted to short stories in English. Although this genre is central to many benchmark datasets, our findings may not generalize to other literary forms, such as poetry, drama, or long-form fiction, which exhibit different stylistic conventions and reader expectations. Genre-specific dynamics remain an open question for future work.

Second, all evaluations are conducted on pre-existing datasets involving blind annotations. While this controls for bias introduced by authorship awareness, it also limits our capacity to account for extratextual variables—such as prior exposure to AI-generated writing, cultural background, or ideological alignment, that may influence reader preferences. These factors likely play a role in real-world literary judgment but lie beyond the scope of this study.

Third, we rely on a curated set of textual features designed to capture stylistic, structural, and semantic properties. Despite efforts to ensure coverage and non-redundancy (Appendix~\ref{appx:correlation}), some relevant dimensions of literary quality—such as narrative engagement, intertextuality, or affective resonance—remain challenging to operationalize. 

Fourth, our preference models are trained on observational data. While they reveal systematic associations between reader types and feature weights, they do not establish causality. Experimental studies manipulating specific features while holding others constant would be needed to isolate causal effects.

Finally, although our clustering analysis identifies robust patterns aligned with reader expertise, the distinction between "expert" and "non-expert" is approximate and dataset-dependent. We define experts as literary professionals (critics, authors, professors), but this category may mask relevant differences in training, taste, or ideology. A finer-grained typology of evaluators is a promising direction for future work.

%By acknowledging these limitations, we aim to clarify the boundaries of our findings and encourage future research to extend this framework across genres, cultures, and reader communities.

%% file: sections/Z_acknowledgements.tex
\section*{Acknowledgements}
This work is partially funded by the Spanish Ministry of Science, Innovation and Universities (project FairTransNLP PID2021-124361OB-C32) funded by MCIN/AEI/10.13039/501100011033, and by a research agreement between UNED and Red.es (C039/21-OT-AD2), which also funded Guillermo Marco's research.

Guillermo Marco's work was further supported by the Spanish government Ph.D. research grant ({\it Ministerio de Universidades}) FPU20/07321 and a scholarship of the Madrid City Council for the Residencia de Estudiantes (Course 2023--2024).

%% file: sections/X_1_all_metrics.tex
%========================================
%  APPENDIX A   Metric Definitions
%  (requires amsmath, amssymb, natbib, mathtools)
%========================================
\section{Formal Definition of the Evaluation Metrics}
\label{app:metrics}

This appendix gives the exact formulae used for every metric reported in the paper.
Throughout, let the evaluated text $x$ consist of:
\begin{itemize}
    \item $S$ sentences $s_1,\dots,s_S$, where $s_i$ has length (number of tokens) $\ell_i$. Each sentence $s_i$ can also be represented by a sentence-level embedding $\mathbf{e}_i\in\mathbb{R}^{d_e}$. The total number of tokens in $x$ is $N = \sum \ell_i$.
    \item $M$ equally sized word-sections $\sigma_1,\dots,\sigma_M$ (typically derived from $x$ by splitting the token sequence) used for global coherence analysis. Each section $\sigma_j$ is associated with a topic vector $\mathbf{t}_j\in\Delta^{K^\star-1}$ derived from an LDA model with $K^\star$ topics.
    \item $K$ fixed-length chunks $c_1,\dots,c_K$ (also derived from $x$ by splitting the token sequence, possibly different from sections $\sigma_j$) used for affect analysis. Each chunk $c_k$ has an associated sentiment score $s_k$ and a dominant emotion $D_k$.
    \item $E_j$ as the set of entities in sentence $s_j$, each entity $e \in E_j$ having a grammatical role $\operatorname{role}_j(e)\!\in\!\{S,O,X\}$ (Subject, Object, Other).
\end{itemize}
For stylistic analysis, $\mathcal{R}$ denotes the predefined set of rhetorical device types considered. For thematic analysis, $C$ is the number of thematic clusters discovered from text segments, and $p_c$ is the proportion of segments belonging to cluster $c$.

{\fontsize{9pt}{12pt}\selectfont  % Apply \small to the entire metrics description block

%-------------------------------------------------
\subsection{Coherence Metrics}
%-------------------------------------------------

\textbf{Optimal number of topics ($K^\star$).}
To determine the optimal number of topics for LDA-based global coherence, we follow \citet{roder2015topic} and select the topic count $K^\star$ that maximizes a topic coherence score, such as $C_v$:
\begin{equation}
  K^\star=\arg\max_{K\in[K_{\min},K_{\max}]} C_{v}(K).
\end{equation}
This $K^\star$ is then used to generate topic vectors $\mathbf{t}_j$ for each section $\sigma_j$.

\textbf{Entity–grid local coherence ($C_{\text{entity}}$).}
Adapting the entity-grid model of \citet{barzilay2008entity}, local coherence is assessed by measuring the continuity of entities and their grammatical roles across adjacent sentences:
\begin{equation}
\begin{split}
  C_{\text{entity}}
     &=\frac{1}{S-1}\sum_{j=1}^{S-1}
        \frac{1}{|E_j\cup E_{j+1}|}
        \sum_{e\in E_j\cup E_{j+1}} \\ % Breakpoint
     &\quad \cdot \Bigl[\mathbf{1}_{e\in E_j\cap E_{j+1}}
        +\tfrac12\,\mathbf{1}_{\substack{e\in E_j\cap E_{j+1} \\ \operatorname{role}_j(e)=\operatorname{role}_{j+1}(e)}}\Bigr].
\end{split}
\label{eq:entity_coherence}
\end{equation}
A higher $C_{\text{entity}}$ suggests smoother transitions in entity focus.

\textbf{Embedding-based local coherence ($C_{\text{local}}$).}
This metric, inspired by \citet{li2014coherence}, quantifies local coherence as the average cosine similarity between the embeddings of adjacent sentences:
\begin{equation}
  C_{\text{local}}=\frac{1}{S-1}\sum_{i=1}^{S-1}\cos(\mathbf{e}_i,\mathbf{e}_{i+1}).
\end{equation}
Higher values indicate greater semantic similarity between consecutive sentences.

\textbf{Global topic-vector coherence ($C_{\text{global}}$).}
Using section-level topic vectors $\mathbf{t}_j$ derived from an LDA model \citep{blei2003lda} with $K^\star$ topics, global coherence is the average cosine similarity across all pairs of section topic vectors:
\begin{equation}
  C_{\text{global}}=\frac{2}{M(M-1)}
    \sum_{1\le i<j\le M}\cos(\mathbf{t}_i,\mathbf{t}_j).
\end{equation}
This captures the overall thematic consistency of the text.

%-------------------------------------------------
\subsection{Originality Metrics}
%-------------------------------------------------
\paragraph{Log-likelihood originality ($\operatorname{Orig}_{\text{LL}}$).}
The originality of text $x$ is assessed by its   log-likelihood under a pre-trained left-to-right language model. Specifically, we utilize the \texttt{meta-llama/Llama-3.2-3B-Instruct} model from the Llama 3 family \citep{aiatmeta2024llama3models}. A higher score, as defined by \citet{brown2020language}, indicates that the text is more predictable by the model, thus considered less original:
\begin{equation}
  \operatorname{Orig}_{\text{LL}}(x)=
    \sum_{n=1}^{N}\log p(w_n\mid w_{<n}),
\end{equation}
where $w_n$ is the $n$-th token and $w_{<n}$ are the preceding tokens.

%-------------------------------------------------
\subsection{Readability Metrics}
%-------------------------------------------------
\textbf{SMOG Index.}
The SMOG grade, developed by \citet{mclaughlin1969smog}, estimates the years of education needed to comprehend a piece of text. It is calculated as:
\begin{equation}
  \operatorname{SMOG}=1.0430\,\sqrt{P \cdot \frac{30}{S_{\text{SMOG}}}}+3.1291,
\end{equation}
where $P$ is the count of polysyllabic words (words with 3 or more syllables) in a sample of 30 sentences, and $S_{\text{SMOG}}$ is the number of sentences in that sample (typically 30, if available).

%-------------------------------------------------
%-------------------------------------------------
\subsection{Sentiment Dynamics Metrics}
%-------------------------------------------------
These metrics analyze the emotional trajectory of the text using chunk-level sentiment scores $s_k$ and dominant emotions $D_k$. Sentiment scores are derived using a fine-tuned RoBERTa-large model for sentiment analysis \citep{hartmann2023more}, while dominant emotions are identified using a RoBERTa-base model fine-tuned on the GoEmotions dataset \citep{demszky2020goemotions, liu2019roberta}.

\textbf{Average sentiment ($\bar{s}$) and Variance ($\sigma_s^2$).}
The overall sentiment tone and its consistency are captured by:
\begin{equation}
\begin{aligned} % 'aligned' allows alignment and stacks lines
  \bar{s} &= \frac{1}{K}\sum_{k=1}^{K}s_k, \\
  \sigma_s^2 &= \frac{1}{K}\sum_{k=1}^{K}(s_k-\bar{s})^{2}.
\end{aligned}
\end{equation}

\textbf{Emotional Volatility (Dominant Emotion Change, $V_{\text{emo-dom}}$).}
This measures the frequency with which the dominant emotion $D_k$ changes between consecutive text chunks:
\begin{equation}
    V_{\text{emo-dom}} = \frac{1}{K-1} \sum_{k=1}^{K-1} \mathbf{1}_{D_k \neq D_{k+1}}.
\end{equation}
A higher value indicates more frequent shifts in the primary emotion conveyed.
% Note: The draft had V_emo for sentiment score changes. Your selected_features has 'emotional_volatility', which points to dominant emotion change in your code.
% If you also want the sentiment score volatility from the draft (mean absolute difference of scores):
% \textbf{Sentiment Score Volatility ($V_{\text{senti-score}}$).}
% The magnitude of change in sentiment scores between adjacent chunks:
% \begin{equation}
%   V_{\text{senti-score}} =\frac{1}{K-1}\sum_{k=1}^{K-1}|s_{k+1}-s_k|.
% \end{equation}

%-------------------------------------------------
\subsection{Stylistic Features}
%-------------------------------------------------
A range of metrics capture stylistic variation and complexity.

\begin{align}
  \shortintertext{Average sentence length (tokens, $\bar{\ell}$):}
      \bar{\ell} &= \frac{1}{S}\sum_{i=1}^{S}\ell_i. \\
  \shortintertext{Average syntactic tree depth ($\bar{d}_{\text{syn}}$), following \citet{lu2010syntactic}:}
      \bar{d}_{\text{syn}} &= \frac{1}{S}\sum_{i=1}^{S}d_{\text{syn}}(s_i), \\
      \label{eq:avg_tree_depth}
      \shortintertext{where $d_{\text{syn}}(s_i)$ is the depth of the syntactic parse tree for sentence $s_i$.} \nonumber \\
  \shortintertext{Lexical Diversity (MTLD), by \citet{mccarthy2010mtld}:}
      \operatorname{MTLD} &= \frac{N}{Q_{0.72}}, \\
      \shortintertext{where $N$ is total tokens and $Q_{0.72}$ is the number of text factors maintaining a Type-Token Ratio $\ge0.72$.} \nonumber \\
  \shortintertext{Maximum subordination depth ($d_{\text{sub-max}}$):}
      d_{\text{sub-max}} &= \max_{i \in \{1,\dots,S\}} \operatorname{sub\_depth}(s_i), \\
      \shortintertext{where $\operatorname{sub\_depth}(s_i)$ is the deepest level of clausal subordination within sentence $s_i$.} \nonumber \\
  \shortintertext{Sentence length standard deviation ($\sigma_\ell$):}
      \sigma_\ell &= \sqrt{\frac{1}{S}\sum_{i=1}^{S}(\ell_i-\bar{\ell})^{2}}. \\
  \shortintertext{Rhetorical device variety ($V_{\text{rhet}}$):}
      V_{\text{rhet}} &= |\{r \in \mathcal{R} \mid \operatorname{count}(r, x) > 0 \}|,
      \label{eq:rhet_variety} \\
      \shortintertext{representing the number of unique types of rhetorical devices from set $\mathcal{R}$ detected in text $x$. The identification of specific devices, such as metaphors, leverages advanced language models (e.g., GPT-4o). This approach aligns with recent findings on the capabilities of Large Language Models for extracting metaphoric analogies, as explored by \citet{boisson2025automatic}.} \nonumber
\end{align}

%-------------------------------------------------
\subsection{Thematic Structure Metrics}
%-------------------------------------------------
These metrics assess the organization and diversity of themes, derived from $C$ thematic clusters found among text segments.

\textbf{Thematic entropy ($H_{\text{themes}}$).}
Normalized entropy of the distribution of text segments across $C$ themes, adapting Shannon's entropy \citep{shannon1948communication}:
\begin{equation}
  H_{\text{themes}} =
     -\frac{1}{\log C}\sum_{c=1}^{C}p_c\log p_c, \quad \text{for } C > 1.
\end{equation}
$H_{\text{themes}}=0$ if $C \le 1$. $p_c$ is the proportion of segments in theme $c$.

\textbf{Thematic graph density ($D_{\text{graph}}$).}
Density of a graph where nodes are themes and edges represent inter-theme semantic similarity above a threshold, based on network theory \citep{newman2010networks}:
\begin{equation}
  D_{\text{graph}} =
     \frac{2E_{\text{th}}}{V_{\text{th}}(V_{\text{th}}-1)}, \quad \text{for } V_{\text{th}} > 1.
\end{equation}
$D_{\text{graph}}=0$ if $V_{\text{th}} \le 1$. $V_{\text{th}}$ is the number of thematic clusters (nodes) and $E_{\text{th}}$ is the number of edges.

\vspace{1em} % Added a bit more space before Notation
\textbf{Notation Summary.}
$\cos(\cdot,\cdot)$ denotes the cosine similarity.
$\mathbf{1}_{\{\text{condition}\}}$ is an indicator function, equal to 1 if the condition is true, and 0 otherwise.
All means and ratios are undefined (or typically set to 0 or NaN, as specified in context) if the denominator is zero.
$\Delta^{k-1}$ is the $(k-1)$-simplex representing probability distributions over $k$ categories.
Unless specified (e.g., SMOG), $S$ refers to the total number of sentences in text $x$.
Logarithms are natural $(\ln)$ unless a base is specified (e.g., $\log_2$). For $H_{\text{themes}}$, the base of the log in the sum and in the normalization factor $1/\log C$ should be consistent (e.g., both natural log, or both base 2). Implementations often use natural log.
} % End of \small block

\input{tables/X_2_metrics}

%% file: tables/X_2_metrics.tex
\begin{table*} % O considera usar el paquete 'rotating' para sidewaystable
    \centering
    \caption{Mean Feature Values per Model/Author}
    \label{tab:dataset_means}
    % --- AJUSTES EXTREMOS PARA ENCAJAR LA TABLA ---
    \tiny % El tamaño más pequeño.
    \setlength{\tabcolsep}{1pt} % Espacio mínimo entre columnas. Prueba 1pt a 2pt.
    \renewcommand{\arraystretch}{0.85} % Reduce la altura de las filas. 0.8-0.9
    % --- FIN DE AJUSTES ---
    \begin{tabular}{llrrrrrrrrrrrrrrrrr}
    \toprule
    Dataset & Author & \makecell[t]{Topic\\count} & \makecell[t]{Entity\\coherence} & \makecell[t]{Local\\coherence} & \makecell[t]{Global\\coherence} & \makecell[t]{Log\\likelihood} & \makecell[t]{SMOG} & \makecell[t]{Mean\\sentiment} & \makecell[t]{Emotional\\swings} & \makecell[t]{Sentiment\\variance} & \makecell[t]{Sentence\\length} & \makecell[t]{Syntactic\\depth} & \makecell[t]{MTLD} & \makecell[t]{Max\\subord...} & \makecell[t]{Rhetorical\\variety} & \makecell[t]{Sentence\\rhythm} & \makecell[t]{Topic\\entropy} & \makecell[t]{Topic\\density} \\
    \midrule % Un midrule después de los encabezados
\multirow{12}{*}{confederacy} & Human & 6.200 & 0.079 & 0.267 & 0.258 & 14323.5 & 8.120 & -0.273 & 0.343 & 0.849 & 12.610 & 2.068 & 115.101 & 3.200 & 12.000 & 9.475 & 0.893 & 0.093 \\
 & alpaca & 7.600 & 0.134 & 0.373 & 0.146 & 1961.4 & 10.900 & 0.928 & 0.507 & 0.100 & 21.967 & 2.012 & -1102.454 & 1.600 & 3.800 & 3.330 & 0.325 & 0.200 \\
 & bard & 6.600 & 0.104 & 0.384 & 0.276 & 8551.8 & 9.020 & 0.202 & 0.346 & 0.887 & 9.712 & 1.852 & 49.805 & 1.800 & 5.200 & 6.344 & 0.797 & 0.467 \\
 & bing & 4.200 & 0.120 & 0.366 & 0.323 & 15107.5 & 8.760 & -0.079 & 0.407 & 0.878 & 11.349 & 2.018 & 87.696 & 2.800 & 10.000 & 7.397 & 0.805 & 0.203 \\
 & chatgpt & 4.700 & 0.037 & 0.336 & 0.327 & 13915.8 & 10.940 & 0.419 & 0.300 & 0.753 & 19.058 & 2.238 & 110.199 & 3 & 9.000 & 12.265 & 0.665 & 0.477 \\
 & claude12 & 5 & 0.033 & 0.314 & 0.358 & 6266.1 & 10.660 & -0.049 & 0.243 & 0.876 & 15.158 & 2.045 & 199.570 & 2.400 & 9.000 & 10.270 & 0.573 & 0.167 \\
 & dolly & 7.200 & 0.064 & 0.295 & 0.141 & 6801.5 & 9.520 & -0.011 & 0.322 & 0.948 & 15.714 & 2.005 & -250.798 & 2.800 & 7.000 & 10.442 & 0.863 & 0.176 \\
 & gpt4all & 5.400 & 0.044 & 0.284 & 0.291 & 7222.5 & 9.180 & 0.779 & 0.365 & 0.361 & 18.661 & 2.112 & 75.783 & 2.800 & 6.800 & 11.693 & 0.606 & 0.400 \\
 & koala & 7.400 & 0.050 & 0.387 & 0.223 & 7631.8 & 10.660 & 0.528 & 0.382 & 0.657 & 18.087 & 2.110 & 92.829 & 3 & 5.600 & 10.226 & 0.736 & 0.207 \\
 & oa & 4.400 & 0.028 & 0.317 & 0.349 & 9177.3 & 10.340 & 0.575 & 0.344 & 0.567 & 17.092 & 2.159 & -3363.100 & 3.200 & 7.400 & 11.562 & 0.863 & 0.279 \\
 & stablelm & 5.800 & 0.056 & 0.377 & 0.266 & 5699.7 & 9.340 & 0.728 & 0.389 & 0.406 & 19.741 & 2.012 & 74.710 & 2.600 & 7.400 & 11.189 & 0.678 & 0.373 \\
 & vicuna & 5.800 & 0.043 & 0.312 & 0.260 & 10636.3 & 9.680 & 0.587 & 0.362 & 0.599 & 17.911 & 2.102 & 72.796 & 2.800 & 7.200 & 11.065 & 0.722 & 0.293 \\
\midrule
\multirow{11}{*}{hanna} & Human & 5.740 & 0.122 & 0.296 & 0.267 & 6960.7 & 7.008 & -0.125 & 0.441 & 0.838 & 16.004 & 1.949 & 47.761 & 2.948 & 8.406 & 9.158 & 0.890 & 0.107 \\
 & bertgeneration & 5.042 & 0.170 & 0.310 & 0.340 & 2747.2 & 6.326 & -0.221 & 0.472 & 0.791 & 14.213 & 1.926 & 88.203 & 2.656 & 5.448 & 8.219 & 0.918 & 0.062 \\
 & ctrl & 5.781 & 0.181 & 0.312 & 0.294 & 2554.5 & 6.014 & -0.239 & 0.474 & 0.752 & 14.466 & 1.849 & -282.722 & 2.677 & 2.802 & 9.036 & 0.918 & 0.059 \\
 & fusion & 6.823 & 0.282 & 0.360 & 0.247 & 1629.3 & 5.071 & -0.130 & 0.476 & 0.706 & 14.135 & 1.862 & 41.879 & 2.281 & 2.312 & 7.147 & 0.740 & 0.031 \\
 & gpt & 5.833 & 0.165 & 0.296 & 0.306 & 2620.6 & 6.204 & -0.191 & 0.484 & 0.832 & 13.345 & 1.897 & -37.285 & 2.531 & 4.583 & 7.448 & 0.928 & 0.074 \\
 & gpt-2 & 4.865 & 0.174 & 0.302 & 0.349 & 4452.7 & 6.511 & -0.157 & 0.497 & 0.839 & 13.343 & 1.888 & 79.343 & 2.792 & 6.729 & 7.563 & 0.902 & 0.077 \\
 & gpt-2 (tag) & 4.948 & 0.160 & 0.306 & 0.337 & 4095.1 & 6.466 & -0.122 & 0.488 & 0.802 & 14.577 & 1.936 & 80.103 & 2.948 & 5.865 & 8.072 & 0.903 & 0.095 \\
 & hint & 5.698 & 0.567 & 0.510 & 0.383 & 1341.9 & 4.438 & -0.133 & 0.405 & 0.548 & 11.824 & 1.883 & 21.624 & 1.469 & 2.031 & 6.403 & 0.374 & 0.038 \\
 & roberta & 5.760 & 0.143 & 0.276 & 0.275 & 2569.8 & 6.578 & -0.207 & 0.478 & 0.789 & 14.901 & 1.954 & 66.908 & 2.510 & 4.719 & 8.236 & 0.939 & 0.053 \\
 & td-vae & 4.375 & 0.072 & 0.309 & 0.356 & 4268.4 & 6.819 & -0.182 & 0.413 & 0.833 & 14.618 & 1.862 & -825.543 & 3.021 & 5.260 & 9.857 & 0.927 & 0.091 \\
 & xlnet & 5.146 & 0.224 & 0.302 & 0.356 & 3679.1 & 6.339 & -0.038 & 0.430 & 0.844 & 14.145 & 1.967 & 50.383 & 2.667 & 3.385 & 8.071 & 0.838 & 0.180 \\
\midrule
\multirow{2}{*}{pronvsprompt} & Human & 4.733 & 0.084 & 0.324 & 0.326 & 7452.0 & 12.967 & 0.054 & 0.283 & 0.908 & 34.585 & 2.513 & 110.984 & 3.300 & 7.983 & 24.497 & 0.883 & 0.092 \\
 & gpt4 & 4.383 & 0.038 & 0.476 & 0.359 & 10557.3 & 13.846 & 0.777 & 0.337 & 0.360 & 25.966 & 2.700 & 99.520 & 3.050 & 8.233 & 16.909 & 0.566 & 0.390 \\
\midrule
\multirow{2}{*}{slm} & Human & 5.148 & 0.095 & 0.263 & 0.188 & 1059.5 & 6.307 & 0.222 & 0.260 & 0.621 & 24.996 & 2.239 & -646.323 & 1.754 & 1.705 & 9.807 & 0.692 & 0.049 \\
 & bart-large & 7.098 & 0.146 & 0.283 & 0.147 & 1113.8 & 3.803 & 0.555 & 0.286 & 0.478 & 28.223 & 2.846 & -239.049 & 2.033 & 0.656 & 6.793 & 0.642 & 0.077 \\
\midrule
\multirow{4}{*}{ttcw} & Human & 7.083 & 0.121 & 0.306 & 0.203 & 20185.0 & 7.575 & -0.119 & 0.303 & 0.911 & 16.204 & 1.992 & 93.312 & 3.583 & 10.500 & 11.016 & 0.924 & 0.046 \\
 & claude & 7.833 & 0.063 & 0.353 & 0.188 & 21671.3 & 8.950 & 0.300 & 0.498 & 0.815 & 16.386 & 2.190 & 132.927 & 3.333 & 9.833 & 9.095 & 0.741 & 0.183 \\
 & gpt3.5 & 8.500 & 0.048 & 0.414 & 0.179 & 33207.6 & 10.083 & 0.741 & 0.603 & 0.427 & 16.412 & 2.216 & 99.939 & 3.500 & 10.333 & 10.276 & 0.509 & 0.285 \\
 & gpt4 & 7.750 & 0.049 & 0.356 & 0.188 & 25545.2 & 10.658 & 0.225 & 0.356 & 0.891 & 21.918 & 2.336 & 130.641 & 3.417 & 11.000 & 15.114 & 0.729 & 0.142 \\
    \bottomrule
    \end{tabular}
\end{table*}

% == PAQUETES LaTeX REQUERIDOS en tu preámbulo ==
% \usepackage{booktabs}
% \usepackage{makecell}
% \usepackage{multirow}
% \usepackage{graphicx}
% \usepackage{amsmath}
% \usepackage{geometry} % e.g., \usepackage[margin=0.5in]{geometry} o landscape
% \usepackage{rotating} % Para \begin{sidewaystable} ... \end{sidewaystable}
% =============================================

%% file: sections/X_3_correlation_analysis.tex
\section{Random Forest Model Performances and Metrics Correlation}
\label{appx:correlation}

\begin{figure*}
    \centering
    \includegraphics[width=0.7\textwidth]{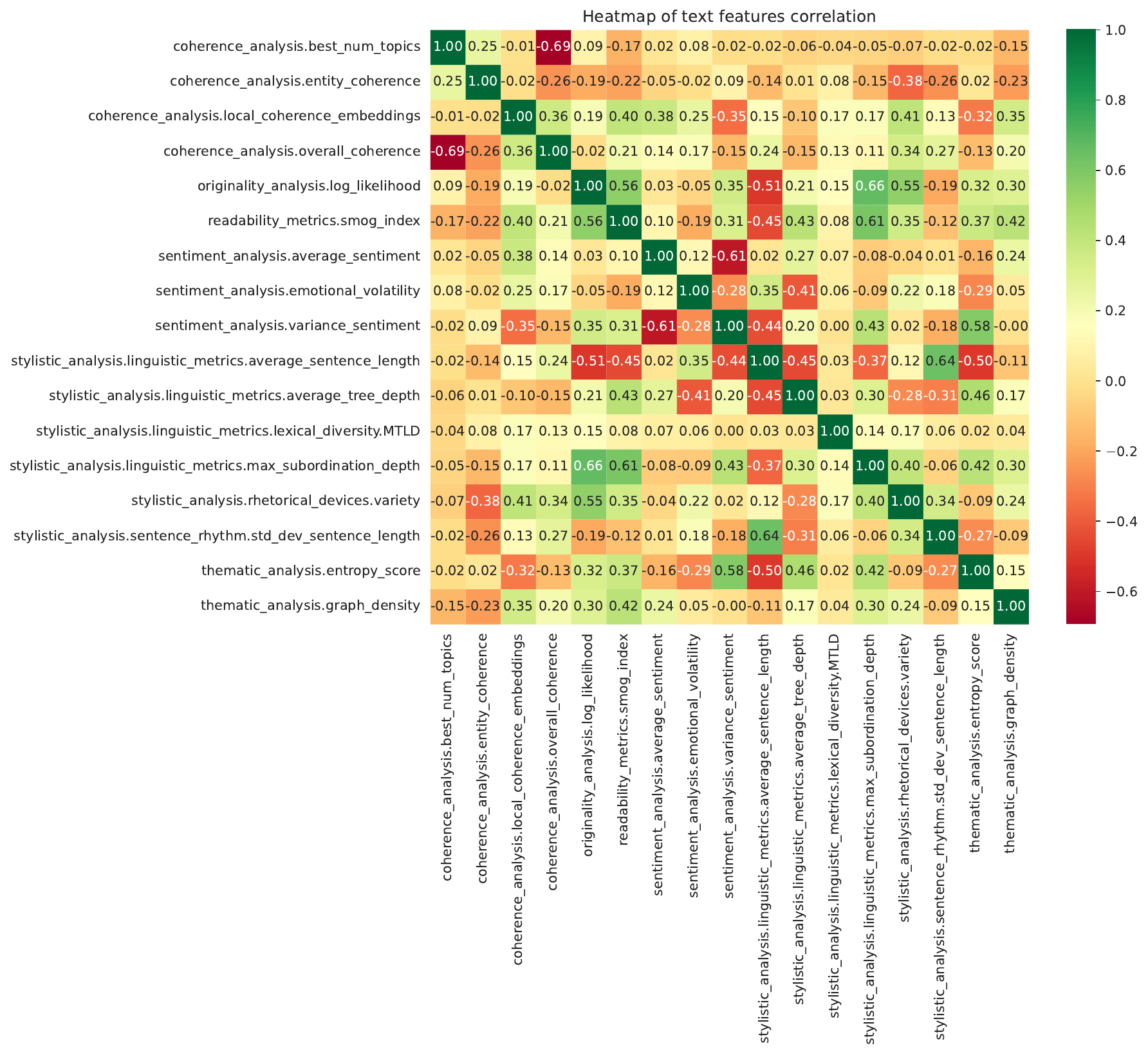}
    \caption{Correlation matrix of the textual features developed across all corpora. Notably, no highly positive or negative correlations are present, ensuring that feature importance is not skewed by redundancy in the Random Forest model.}
    \label{fig:correlations}
\end{figure*}

This section addresses the validation of the selected metrics, models, and feature importances. Given the sensitivity of Random Forest feature importances with Gini entropy in distributing importance among correlated variables, we provide the Figure~\ref{fig:correlations}. This table demonstrates that none of the variables used to represent the texts and train the models exhibit correlations higher than 0.7.

\begin{figure*}[htbp]
    \centering
    \begin{subfigure}[b]{0.4\textwidth} % Más pequeñas
        \centering
        \includegraphics[width=\linewidth]{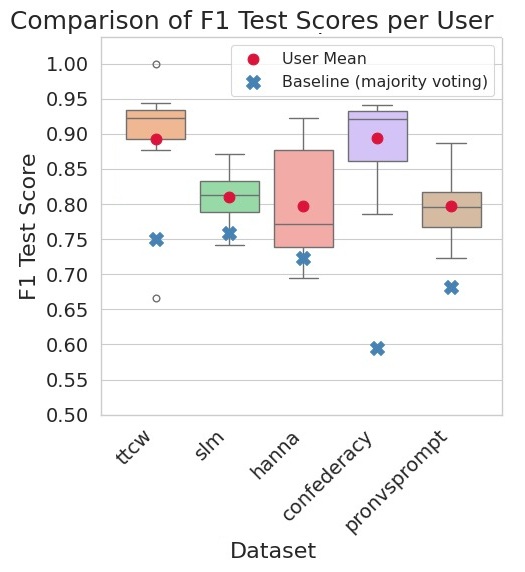}
        \caption{F1 scores on test set - Random Forest}
        \label{fig:random_forest_f1}
    \end{subfigure}
    \hspace{0.05\textwidth} % Un espacio horizontal fijo (5% del ancho del texto)
    \begin{subfigure}[b]{0.4\textwidth} % Más pequeñas
        \centering
        \includegraphics[width=\linewidth]{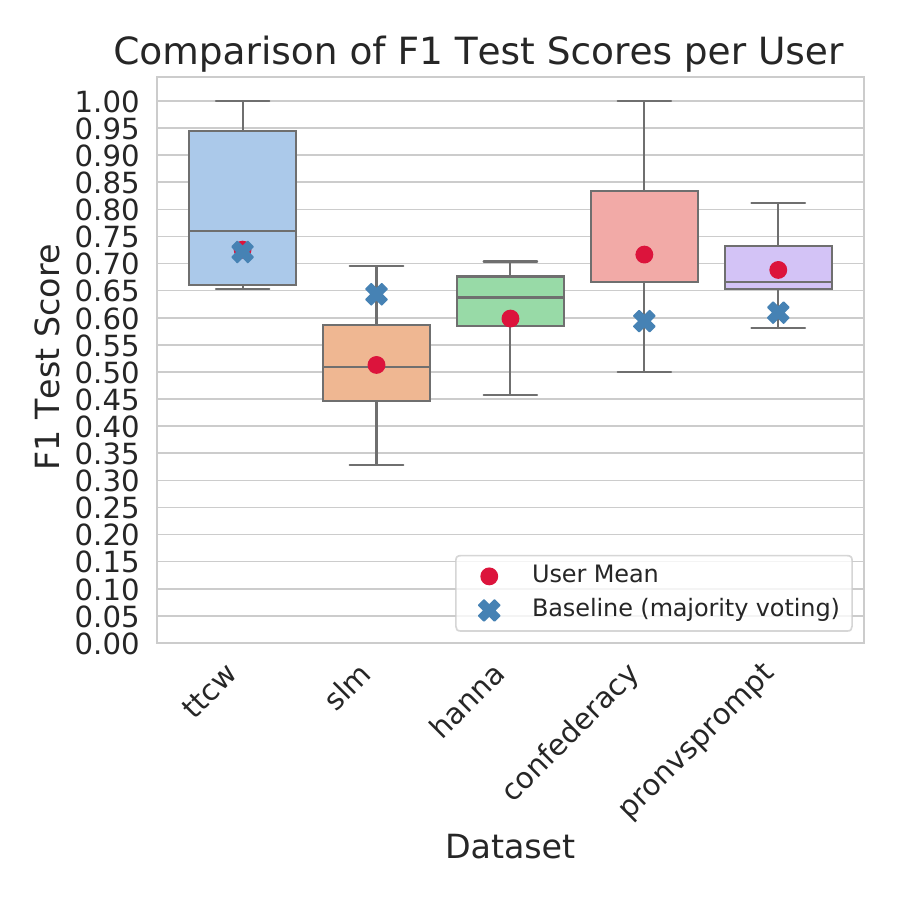}
        \caption{F1 scores on test set - Logistic Regression}
        \label{fig:logistic_regression_f1}
    \end{subfigure}
    \caption{Comparison of F1 scores on test set for Random Forest and Logistic Regression. The boxplots illustrate the distribution of F1 scores across different datasets when modeling user preferences individually. The red dots indicate the mean F1 score for each user, while the blue crosses represent the performance of the baseline model trained on text pairs with majority voting consensus.}
    \label{fig:f1}
\end{figure*}

Additionally, Figure \ref{fig:f1} presents the visualization of F1 scores on the test set, illustrating the model’s ability to predict preferences for previously unseen text pairs. Hyperparameters for both were tuned via grid search. For Logistic Regression, we explored $C \in \{0.01, 0.1, 1, 10\}$ with L1/L2 regularization. For Random Forest, parameters included $n\_estimators \in \{100, 200, 300\}$, $max\_depth \in \{None, 10, 20, 30\}$, $min\_samples\_split \in \{2, 5, 10\}$, $min\_samples\_leaf \in \{1, 2, 4\}$, and bootstrap/non-bootstrap sampling. Random Forest outperformed Logistic Regression, achieving higher F1 scores on the test set.

%% file: sections/X_2_dataset_limitations.tex
\begin{table*}
\scriptsize
\section{Dataset Profiles}
\subsection*{C.1 Dataset Profile: SLM \cite{marco_small_2025}}

\begin{minipage}{\textwidth}
\begin{tabularx}{\textwidth}{@{}p{4cm} X@{}}
\toprule
\textbf{Text genre and format} &
Fictional movie synopses (approximately 80 words), written in response to a title. Half of the texts are human-written (collected and lightly edited from public film databases such as Wikipedia and TMDb), and the other half are generated by a fine-tuned BART-large model. \\
\midrule
\textbf{Genre homogeneity} &
All texts conform to the same narrative subgenre—short, single-paragraph movie synopses. \\
\midrule
\textbf{Corpus size} &
60 texts per condition (30 human-written and 30 AI-generated), evaluated under three experimental framings: (i) neutral (no authorship information), (ii) with explicit authorship labels, and (iii) deceptive framing (all texts labeled as AI regardless of source). In total, over 24,000 Likert-scale ratings were collected from human evaluators. \\
\midrule
\textbf{Annotator profile} &
68 master's students from an international business program. Participants had no formal literary training but were diverse in academic background and nationality. The sample is representative of a general, educated readership rather than literary experts. \\
\midrule
\textbf{Evaluation design} &
A between-subjects design was used: each participant rated only one synopsis per title, with balanced assignment to human or AI variants. The framing condition (authorship label) was randomized between participants. Texts were rated on multiple dimensions including creativity, informativeness, and overall quality. Two quiz variants were designed to balance exposure and ensure coverage. \\
\midrule
\textbf{Strengths} &
\begin{itemize}[topsep=0pt, partopsep=0pt, itemsep=0pt, parsep=0pt] 
    \item Large number of ratings allows for robust statistical modeling of individual and aggregate preferences.
    \item The use of framing conditions enables analysis of expectation bias in judgments of creativity.
    \item Uniform text format enhances internal validity and minimizes stylistic confounds.
\end{itemize} \\
\midrule
\textbf{Limitations} &
\begin{itemize}[topsep=0pt, partopsep=0pt, itemsep=0pt, parsep=0pt] 
    \item The genre is highly constrained (film synopses only), which may not reflect broader creative writing capabilities.
    \item Human-written texts were not authored by professional writers and may not serve as high literary benchmarks.
    \item All annotators were non-experts, potentially biasing evaluations toward surface-level qualities (e.g., clarity, coherence).
\end{itemize} \\
\midrule
\textbf{Observations} &
\begin{itemize}[topsep=0pt, partopsep=0pt, itemsep=0pt, parsep=0pt] 
    \item AI-generated synopses outperformed human-written ones on most dimensions, except for creativity, where human texts had a slight edge.
    \item Framing had a significant impact: revealing AI authorship led to lower ratings, while falsely labeling human texts as AI increased their perceived quality.
    \item Creativity scores correlated strongly with informativeness, but not with attractiveness, suggesting distinct evaluative dimensions.
\end{itemize} \\
\bottomrule
\end{tabularx}
\end{minipage}
\caption{}
\label{tab:dataset_slm}
\end{table*}

\begin{table*}
\scriptsize
\subsection*{C.2 Dataset Profile: Confederacy \cite{gomez-rodriguez_confederacy_2023}}

\begin{minipage}{\textwidth}
\begin{tabularx}{\textwidth}{@{}p{3.8cm} X@{}} % Reducido ligeramente el ancho de la primera columna
\toprule
\textbf{Text genre \& format} & % "&" en lugar de "and"
65 short stories (250-1200 words) from a single prompt: “Write an epic narration of a single combat between Ignatius J. Reilly and a pterodactyl, in the style of John Kennedy Toole.” 5 human-authored, 60 AI-generated (5 per model from 12 LLMs). \\
\midrule
\textbf{Genre homogeneity} &
Extremely high. All stories from the same prompt, targeting humor, epic structure, and Toole's style. Ensures comparability but limits generalization. \\
\midrule
\textbf{Corpus size} &
65 texts (5 human, 60 AI). Each rated by 2 independent raters on 10 dimensions (1300 ratings total), plus qualitative commentary. Each rater reviewed 13 texts. \\
\midrule
\textbf{Annotator profile} &
10 Honours/postgraduate Creative Writing students. Annotations blind to authorship. \\
\midrule
\textbf{Evaluation design} &
Stories rated 1-10 across 10 dimensions (e.g., plot, humor, style fidelity, originality, coherence). Implicit ranking. Annotators provided qualitative feedback. \\
\midrule
\textbf{Strengths} &
\begin{itemize}[topsep=0pt, partopsep=0pt, itemsep=0pt, parsep=0pt] 
    \item Exceptional control over genre, topic, and style via unified prompt.
    \item Evaluation rubric adapted from creative writing pedagogy.
    \item Direct comparison of multiple SOTA LLMs and human performance.
    \item Public availability of texts, annotations, and metadata.
\end{itemize} \\
\midrule
\textbf{Limitations} &
\begin{itemize}[topsep=0pt, partopsep=0pt, itemsep=0pt, parsep=0pt] 
    \item Small human sample (5 stories), limiting human stylistic variation.
    \item Only two annotators per story; moderate agreement ($k = 0.48$).
    \item Highly specific task; results may not generalize broadly.
    \item Some models (e.g., Bing) affected by prompt censorship, potentially skewing outputs.
\end{itemize} \\
\midrule
\textbf{Observations} &
\begin{itemize}[topsep=0pt, partopsep=0pt, itemsep=0pt, parsep=0pt] 
    \item GPT-4 achieved highest mean scores, outperforming humans in most dimensions.
    \item Humans retained a relative advantage in humor and originality.
    \item Some models produced readable but stylistically flat stories.
    \item LLMs excelled in formal aspects like epicness and structure.
\end{itemize} \\
\bottomrule
\end{tabularx}
\end{minipage}
\caption{}
\label{tab:dataset_confederacy}
\end{table*}

\begin{table*}
\scriptsize
\subsection*{C.3 Dataset Profile: Pron vs Prompt \cite{marco_pron_2024}}

\begin{minipage}{\textwidth}
\begin{tabularx}{\textwidth}{@{}p{3.8cm} X@{}}
\toprule
\textbf{Text genre \& format} &
180 fictional film synopses (~600 words each). 60 by human author Patricio Pron (Spanish). 120 by GPT-4: 60 using its own titles (30 ES, 30 EN) and 60 using Pron's titles (30 ES, 30 EN). \\
\midrule
\textbf{Genre homogeneity} &
High thematic/structural consistency (imaginary film synopses). Stylistic richness varies by title provenance and language, introducing controlled variation. \\
\midrule
\textbf{Corpus size} &
180 stories. 6 expert evaluators rated 120 stories each, yielding 7,200 judgments across creativity-related dimensions. Blind to authorship. \\
\midrule
\textbf{Annotator profile} &
Six literary experts (scholars, editors, instructors). Three assessed Spanish texts (GPT-4 + Pron); three bilingual experts assessed English GPT-4 texts vs. Spanish Pron texts. \\
\midrule
\textbf{Evaluation design} &
Rubric inspired by M. Boden’s creativity dimensions (e.g., originality, narrative voice, attractiveness). Experts guessed authorship (human/AI) and gave qualitative comments. Blind evaluations. \\
\midrule
\textbf{Strengths} &
\begin{itemize}[topsep=0pt, partopsep=0pt, itemsep=0pt, parsep=0pt] 
    \item First direct, controlled comparison between a world-class author and GPT-4.
    \item Expert-driven rubric based on cognitive theory of creativity.
    \item Cross-lingual setup reveals GPT-4 performance asymmetries by language.
    \item Systematic manipulation of prompt provenance (human vs. AI titles) for co-authorship analysis.
\end{itemize} \\
\midrule
\textbf{Limitations} &
\begin{itemize}[topsep=0pt, partopsep=0pt, itemsep=0pt, parsep=0pt] 
    \item Human texts from a single author (limits diversity).
    \item Task-specific (synopses); may not generalize to other genres.
    \item Expert-only evaluations; no general-audience reactions.
    \item Potential influence of language asymmetry in bilingual expert comparisons.
    \item Only one AI model (GPT-4) evaluated.
\end{itemize} \\
\midrule
\textbf{Observations} &
\begin{itemize}[topsep=0pt, partopsep=0pt, itemsep=0pt, parsep=0pt] 
    \item GPT-4’s performance significantly improved with human-authored titles (strong prompt sensitivity).
    \item Experts' AI authorship detection improved over time (consistent GPT-4 stylistic signals).
    \item GPT-4 performed better in English than Spanish across most dimensions.
    \item GPT-4 rarely and inconsistently matched or surpassed the human author.
\end{itemize} \\
\bottomrule
\end{tabularx}
\end{minipage}
\caption{}
\label{tab:dataset_pron}
\end{table*}

\begin{table*}
\scriptsize
\subsection*{C.4 Dataset Profile: TTCW \cite{chakrabarty_art_2024}}

\begin{minipage}{\textwidth}
\begin{tabularx}{\textwidth}{@{}p{3.8cm} X@{}}
\toprule
\textbf{Text genre \& format} &
48 short stories (~1000-2400 words), in 12 thematic quartets. Each: 1 \textit{New Yorker} story (human), 3 AI-generated (GPT-3.5, GPT-4, Claude v1.3) from same one-sentence plot summary. Diverse themes/styles. \\
\midrule
\textbf{Genre homogeneity} &
All short fiction narratives from shared plot origin. Human texts are high-literary; AI attempts stylistic emulation. Ensures comparability, though human texts are from a narrow, high-end domain. \\
\midrule
\textbf{Corpus size} &
48 stories (12 human, 36 AI). Each evaluated by 3 experts using a 14-item binary rubric (adapted TTCT), yielding >2,000 creativity annotations. \\
\midrule
\textbf{Annotator profile} &
10 experts: creative writing professors, literary agents, MFA-trained writers. Each story assessed independently by 3 experts; provided comparative rankings and authorship guesses. \\
\midrule
\textbf{Evaluation design} &
Used adapted Torrance Test of Creative Writing (TTCW): 14 binary (yes/no) assessments in 4 dimensions (Fluency, Flexibility, Originality, Elaboration). Experts gave free-text commentary, rankings, blind authorship attribution. \\
\midrule
\textbf{Strengths} &
\begin{itemize}[topsep=0pt, partopsep=0pt, itemsep=0pt, parsep=0pt] 
    \item First application of a structured creativity test (TTCT) to literary evaluation.
    \item Balanced story lengths and aligned prompts enhance control.
    \item High-quality expert judgments based on an established creativity framework.
    \item Strong inter-rater reliability at aggregate level (Pearson $\rho = 0.69$).
\end{itemize} \\
\midrule
\textbf{Limitations} &
\begin{itemize}[topsep=0pt, partopsep=0pt, itemsep=0pt, parsep=0pt] 
    \item Human texts are high-literary, potentially overstating AI–human gap.
    \item No mid-tier/casual human writing; limited human stylistic diversity.
    \item Evaluation process is time-consuming and hard to scale.
    \item One-sentence plot conditioning may limit AI narrative divergence.
    \item TTCW’s binary rubric might not capture all nuances of creative quality.
\end{itemize} \\
\midrule
\textbf{Observations} &
\begin{itemize}[topsep=0pt, partopsep=0pt, itemsep=0pt, parsep=0pt] 
    \item Human stories passed 84.7\% of TTCW tests; GPT-4 27.9\%, Claude v1.3 30.0\%.
    \item Experts ranked human stories as best in 89\% of quartets.
    \item Claude v1.3 was more often mistaken for human than GPT-based models.
    \item Claude v1.3 performed best across Fluency, Flexibility, Elaboration; GPT-4 highest in Originality.
\end{itemize} \\
\bottomrule
\end{tabularx}
\end{minipage}
\caption{}
\label{tab:dataset_ttcw}
\end{table*}

\begin{table*}
\scriptsize
\subsection*{C.5 Dataset Profile: HANNA \cite{chhun_human_2022}} % He añadido el nombre del dataset

\begin{minipage}{\textwidth}
\begin{tabularx}{\textwidth}{@{}p{3.8cm} X@{}}
\toprule
\textbf{Text genre \& format} &
1,056 stories generated by 10 different Automatic Story Generation (ASG) systems from 96 unique prompts (short sentences from WritingPrompts dataset). Each prompt is linked to a human-written reference story. \\
\midrule
\textbf{Genre homogeneity} &
High input homogeneity (short sentence prompts). Generated stories vary based on diverse prompts and system capabilities. \\
\midrule
\textbf{Corpus size} &
1,056 stories. Each annotated by 3 human raters (MTurk) on 6 criteria (19,008 annotations total). Includes scores from 72 automatic metrics. \\
\midrule
\textbf{Annotator profile} &
Amazon Mechanical Turk workers, filtered for English fluency, location (UK, US, CA, AU, NZ), and Masters Qualification. \\
\midrule
\textbf{Evaluation design} &
6 human criteria: Relevance (RE), Coherence (CH), Empathy (EM), Surprise (SU), Engagement (EG), Complexity (CX), rated on a 5-point Likert scale. Correlation analysis with 72 automatic metrics. \\
\midrule
\textbf{Strengths} &
\begin{itemize}[nosep]
    \item Comprehensive set of 6 orthogonal human evaluation criteria motivated by social sciences.
    \item Large, publicly available dataset (HANNA) with human annotations and automatic scores.
    \item Extensive meta-evaluation comparing 72 automatic metrics against human criteria.
    \item Detailed inter-annotator agreement analysis (ICC2k reported).
\end{itemize} \\
\midrule
\textbf{Limitations} &
\begin{itemize}[nosep]
    \item Inter-annotator agreement is fair to moderate (ICC2k 0.29-0.56).
    \item Annotators are MTurk workers, not literary experts.
    \item Focused on stories generated from WritingPrompts; may not generalize to other ASG tasks/inputs.
    \item Existing automatic metrics show weak correlation with human judgments, especially at story level.
\end{itemize} \\
\midrule
\textbf{Observations} &
\begin{itemize}[nosep]
    \item Human-written stories consistently score higher than AI-generated ones across all criteria.
    \item Among ASG systems, GPT-2 (generic fine-tuned) performed best.
    \item Automatic metrics correlate better with human judgments at system-level than story-level.
    \item chrF and BARTScore are among the better performing automatic metrics, but human annotation is still advised.
\end{itemize} \\
\bottomrule
\end{tabularx}
\end{minipage}
\caption{}
\label{tab:dataset_hanna}
\end{table*}